\documentclass[journal]{IEEEtran}
\usepackage[utf8]{inputenc}
\usepackage{textcomp}
\usepackage{graphicx}
\usepackage[version=4]{mhchem}
\usepackage{siunitx}
\usepackage{longtable,tabularx}
\usepackage{accents}
\usepackage{comment}
\usepackage{mathtools}
\usepackage{subfigure}
\usepackage{caption}
\usepackage{color}
\usepackage{amsmath,amssymb}
\usepackage{booktabs}
\usepackage{tabu}
\usepackage{algorithm}
\usepackage{algpseudocode}
\usepackage{hyperref}
\usepackage{soul}
\usepackage[dvipsnames]{xcolor}
\usepackage{multirow}
\usepackage{enumerate}
\usepackage{svg}
\usepackage{epstopdf}
\usepackage{arydshln}
\usepackage[colorinlistoftodos]{todonotes}
\usepackage{verbatim}
\usepackage[normalem]{ulem}
\usepackage{enumitem}

\usepackage{tikz}
\usepackage{siunitx} 
\usetikzlibrary{arrows, arrows.meta}
\usetikzlibrary{graphs}
\usetikzlibrary{snakes,backgrounds,patterns,matrix,shapes,fit,calc,shadows,plotmarks} 
\usetikzlibrary{bayesnet}
\usetikzlibrary{matrix}
\usetikzlibrary{fit}

\usepackage{standalone}

\tikzstyle{target}=[draw,fill=yellow!50,circle,minimum size=16pt,inner sep=0pt]

\tikzstyle{output}=[draw,fill=blue!50,circle,minimum size=16pt,inner sep=0pt]
\tikzstyle{bias}=[draw,fill=gray!50,circle,minimum size=20pt,inner sep=2pt]
\tikzstyle{arrow}=[arrows={{Latex[scale=0.5]}-}, thick]  % , in=90, out=-90
\tikzstyle{box}=[rectangle, draw=black!100] 

\newcommand{\od}[1]{{\color{ForestGreen}OD: #1}}
\newcommand{\odb}[1]{{\color{blue}#1}}

\newcounter{mycomment}

\newcommand{\infoMat}{\begin{pmatrix}
\Lambda_{\chi_C\chi_C}  &  \Lambda_{\chi_C\chi_L} \\
\Lambda_{\chi_L\chi_C}  & \Lambda_{\chi_L\chi_L} 
\end{pmatrix}}
\newcommand{\covMat}{\begin{pmatrix}
\Sigma_{\chi_C\chi_C}  &  \Sigma_{\chi_C\chi_L}\\
\Sigma_{\chi_L\chi_C}  & \Sigma_{\chi_L\chi_L} 
\end{pmatrix}}

\newcommand{\etal}{\emph{et al. }}

\ifCLASSINFOpdf
\else
\fi
\begin{document}
\title{Exact and Approximate Heterogeneous \\ Bayesian Decentralized Data Fusion}

\author{Ofer~Dagan,~\IEEEmembership{Student Member,~IEEE,}
        Nisar~R.~Ahmed,~\IEEEmembership{Member,~IEEE}% <-this % stops a space
\thanks{Manuscript received 27 May 2022; revised 26 August 2022; accepted 10 November 2022. Date of publication 14 December 2022; date of current version 5 April 2023. This paper was recommended for publication by Associate Editor R. Tron and Editor P. Robuffo Giordano upon evaluation of the reviewers’ comments. (Corresponding author: Ofer Dagan.)}
\thanks{The authors are with the Smead Aerospace Engineering Sciences Department, University of Colorado Boulder, Boulder, CO 80309 USA (e-mail: ofer.dagan@colorado.edu; Nisar.Ahmed@colorado.edu). }
\thanks{ "© 2023 IEEE.  Personal use of this material is permitted.  Permission from IEEE must be obtained for all other uses, in any current or future media, including reprinting/republishing this material for advertising or promotional purposes, creating new collective works, for resale or redistribution to servers or lists, or reuse of any copyrighted component of this work in other works."}
}% <-this % stops a space


\markboth{IEEE Transactions on Robotics ,~Vol.~39, No.~2, APRIL~2023}%
{Shell \MakeLowercase{\textit{et al.}}: Bare Demo of IEEEtran.cls for IEEE Journals}



\maketitle

\begin{abstract}
In Bayesian peer-to-peer decentralized data fusion, the underlying distributions held locally by autonomous agents are frequently assumed to be over the same set of variables (homogeneous). This requires each agent to process and communicate the full global joint distribution, and thus leads to high computation and communication costs irrespective of relevancy to specific local objectives. This work formulates and studies heterogeneous decentralized fusion problems, defined as the set of problems in which either the communicated or the processed distributions describe different, but overlapping, random states of interest that are subsets of a larger full global joint state.
We exploit the conditional independence structure of such problems and provide a rigorous derivation of novel exact and approximate conditionally factorized heterogeneous fusion rules. We further develop a new version of the homogeneous \emph{Channel Filter} algorithm to enable conservative heterogeneous fusion for smoothing and filtering scenarios in dynamic problems. Numerical examples show more than 99.5\% potential communication reduction for heterogeneous channel filter fusion, and a multi-target tracking simulation shows that these methods provide consistent estimates while remaining computationally scalable.
\end{abstract}


% \fi
\IEEEpeerreviewmaketitle

\section{Introduction}
Bayesian decentralized data fusion (DDF) has a wide range of applications, such as cooperative localization \cite{loefgren_scalable_2019}, multi-target tracking \cite{whitacre_decentralized_2011}, multi-robot localization and mapping \cite{cunningham_ddf-sam_2013}, and more.
DDF, while generally less accurate compared to centralized data fusion, offers advantages in terms of scalability, flexibility and robustness. 
One of the challenges of decentralized data fusion stems from the difficulty of accounting for common data and dependencies between communicating agents and avoiding %double counting and 
`rumor propagation', where dependencies between sources are incorrectly ignored, causing data or prior information (e.g. process noise) to be counted more than once. %%dependent information is incorrectly treated as independent information. %previously fused information is incorrectly treated as novel. 

In decentralized multi-agent systems aiming at some joint mission, such as autonomous cooperative robot localization  \cite{loefgren_scalable_2019},\cite{li_cooperative_2012}, \cite{zhu_cooperative_2019}, or tracking \cite{whitacre_decentralized_2011}, the optimal solution can be achieved only if each agent recursively updates and communicates a global full joint posterior probability distribution function (pdf) over an identical (homogeneous) set of random variables (rvs), such that all dependencies in the data can be accounted for. This leads to large overhead in local processing and in communication bandwidth. It is therefore desirable to enable processing, communication and fusion of a posterior pdf over a subset of different but overlapping rvs; we name such a process \emph{heterogeneous fusion}. 
Consider for example the 30 robot cooperative localization scenario given in \cite{loefgren_scalable_2019}. 
If each agent has 4 unknown random position states, the full joint distribution has 120 variables, and requires processing a $120\times 120$ covariance matrix (assuming Gaussian pdf). 
This includes states of agents `far away' from each other in the network, which has negligible effect on the local position estimate and might be considered `irrelevant'.  But, if each agent includes in its estimate a subset of only immediate `relevant' neighbors' states, then the local heterogeneous joint distribution shrinks, e.g. to 16 states for a 3-neighbor topology. This has a clear computation and communication gain over the homogeneous alternative. However it might lead to indirect dependencies between variables not mutually monitored by both agents and result in an overconfident estimate. While there are existing methods that allow for heterogeneous fusion, as detailed later, they are application-specific and do not explore or give general insights for a solution to the full heterogeneous fusion problem, as described in this paper.

DDF algorithms can be considered exact or approximate, depending on how they account for dependencies in the data shared between agents in order to guarantee that every piece of data is counted only once. 
In exact methods, these dependencies are explicitly accounted for either by pedigree-tracking, which can be cumbersome and impractical in large ad hoc networks \cite{martin_distributed_2005}, or by adding a 'Channel Filter' (CF) \cite{grime_data_1994}. Approximate methods assume different levels of dependency between the communicating agents and fuse them in such a way that the common data is promised not to be double counted. This is a necessary condition to ensure conservativeness of the fused posterior pdf, where conservative means that the approximate posterior does not underestimate the uncertainty relative to the true pdf.
The most commonly used approximate method is covariance intersection (CI) \cite{julier_non-divergent_1997}, where agents share only the first two moments (mean and covariance) of their underlying distributions (often representing fusion of estimates), or the geometric mean density (GMD) for general pdf fusion \cite{bailey_conservative_2012}. %In CI, the fused result is a weighted average of the information vector and matrix describing the first and second posterior moments, %of the underlying distributions, 
%respectively, where the weight is optimized based on a predetermined cost function, e.g., determinant of the posterior fused covariance matrix. 
%It is usually assumed in exact and approximate fusion that the posterior distributions to be fused  %as well as the posterior distributions 
%are homogeneous, i.e. describe the same full set of rvs. 
Critically, it is usually assumed in both exact and approximate fusion that the process takes place over homogeneous states, i.e. that all underlying posteriors describe the same full set of rvs. 
Thus, these methods cannot be directly applied to heterogeneous fusion, where pdfs with rvs from different overlapping parts of the full joint pdf are fused. %\nisarComm{
This work's main insight is to understand how conditional independence properties of the problem can be leveraged and maintained in such cases to achieve correct and beneficial fusion at scale. %%(NRA: this needs to be stated here and restated at key points, otherwise reader gets lost in all the other details...)

%\nisarComm{In addressing the AE's comments to shrink paper and compress redundant bits a bit more (since Intro is rather long -- should be reduced to 1.5 pages max): I would suggest trying the following: (1) move paragraph (**) below to this position, i.e. right after the preceding paragraph; (2) merge paragraphs (*) and (***): in (*), you are talking about the general goals/aims and specific assumptions of the paper, and then you switch to talking about other general DDF algorithms and their limitations in (**), and then go back in (***) to talking about your paper and the details again-- this is a logical speedbump and it feels redundant to state certain items in (*) and (***) separately when you can just fuse them together. Otherwise, the way it is structured now makes the reader work harder to follow what you are really saying, and it is not clear what the "single sentence' take home message is (e.g. see end of (**), which I added in -- this was buried and not very well articulated in (***), which is why I think the AE is being fussy. From my read and re-read of his comments: he generally likes what he sees, but can't put his finger on THE sentence in the intro text that explains to the reader up front what all the other words/math is about.}

The goal of this paper is to define and explore the heterogeneous DDF problem, suggest new fusion rules, and understand their limitations. %, i.e. when it should work and when it might fail. %To this aim, a probabilistic Bayesian approach is taken and novel heterogeneous DDF rules are derived. These allow the representation and solution of a variety of multi-agent fusion problems and are not tailored around a specific application. 
The paper builds upon the work presented in \cite{dagan_heterogeneous_2020} and further develops a rigorous Bayesian probabilistic approach for fusion of heterogeneous pdfs.
In developing heterogeneous fusion rules for robotics applications, challenges arise from: (i) non-Gaussianity of the true underlying system; and (ii) dynamics. To gain a better fundamental understanding of the issues and requirements for corresponding fusion rules applicable to robotics,
the analysis is constructed as follows. First, in Sec. \ref{sec:factorCF}, the general probabilistic formulation of the homogeneous fusion rule is described. 
We initially assume a static system and develop two fusion rules by exploiting and maintaining conditional independence properties between subsets of random variables (e.g., states). These fusion rules, namely the \emph{Bi-directional factorized fusion} (BDF-fusion) and the \emph{Heterogeneous state fusion} (HS-fusion), allow the communication of only marginal pdfs corresponding to new and relevant data.   
%The key point of exploiting and maintaining conditional independence properties between subsets of rvs (e.g., states) is made. By initially assuming a static system, we leverage conditional independence to develop two fusion rules, namely the \emph{Bi-directional factorized fusion} (BDF-fusion) and the \emph{Heterogeneous state fusion} (HS-fusion), which allow the communication of only marginal pdfs corresponding to new and relevant data. 
Next, in Sec. \ref{sec:DynamicSystems}, the static assumption is relaxed and the problem of conditional independence in dynamic systems is discussed with its solution approach. 
Finally, Sec. \ref{subSec:Gaussians} completes the theoretical analysis, gives further intuition into the heterogeneous fusion problem, and suggests a practical heterogeneous fusion algorithm (Algorithm \ref{algo:CF2}) by considering Gaussian pdfs in linear dynamic systems and expanding the original (homogeneous) CF algorithm \cite{grime_data_1994} to heterogeneous fusion. While the CF algorithm requires the communication graph to be undirected and acyclic, to ensure that information in the network do not `circle' back to its sender, it provides some fundamental insights as shown in this paper. While it is beyond the scope of this work, these insights enable to rigorously extend ideas and algorithms for heterogeneous DDF to other settings where dependent information flows cannot be explicitly or easily tracked (e.g. for fusion in networks with dynamic/cyclic communication graphs). 
%To complete the theoretical analysis and suggest practical heterogeneous fusion algorithm, the case of Gaussian distributions in linear dynamic systems is given in Sec. \ref{subSec:Gaussians}.

%The key idea is to utilize conditional independence properties between subsets of rvs (e.g., states) in order to lower the dimension of the local joint distributions to be fused. 
%In \cite{dagan_heterogeneous_2020} the applicability of the conditionally factorized channel filter (CF$^2$) algorithms for dynamic systems is limited since the full state time history has to be tracked for certain key conditional independence assumptions needed to perform heterogeneous Bayesian DDF to hold. 

This paper improves and extends the theory and applicability of the work in \cite{dagan_heterogeneous_2020} by making the following contributions: 
\begin{enumerate}
    \item The definition and desiderata for heterogeneous fusion rules are more formally and clearly defined, along with practical measures for evaluation. 
    \item The theory and fusion rules developed in this paper are brought to practice with a detailed explanation of two heterogeneous fusion algorithms, and pseudo-code for the case of Gaussian distributions is presented.
    \item Heterogeneous fusion rules are extended to dynamic problems by
    (i) enabling conservative filtering, and (ii) the linear information augmented state (\emph{iAS}) smoother.
    %\item The scalability of the heterogeneous fusion methods is demonstrated by an extended example, calculating the communication and computation savings, and covering different problem sizes.
    %\item New simulations of dynamic decentralized multi-target tracking scenarios are provided, to demonstrate conservative filtering and compare with the full time history smoothing approach. 
    \item Extended numerical example and simulation studies, demonstrating the heterogeneous fusion methods scalability and applicability for dynamic decentralized multi-target tracking scenarios using conservative filtering.
\end{enumerate}

The rest of this paper is organized as follows: Sec. \ref{sec:ProbStatement} defines the heterogeneous decentralized fusion problem and reviews the state of the art; Sec. \ref{sec:factorCF}-\ref{subSec:Gaussians} develops the theory and heterogeneous fusion rules for static and dynamic systems.
Numerical examples demonstrate the potential communication and computation gains of the described methods (Sec. \ref{sec:NumExample}), and simulation study demonstrates its applicability via static and dynamic multi-agent multi-target tracking scenarios (Sec. \ref{sec:Sim}). 
Sec. \ref{sec:conclusions} then draws conclusions and suggests avenues for additional work.

\section{Problem Statement and Related Work}
\label{sec:ProbStatement}
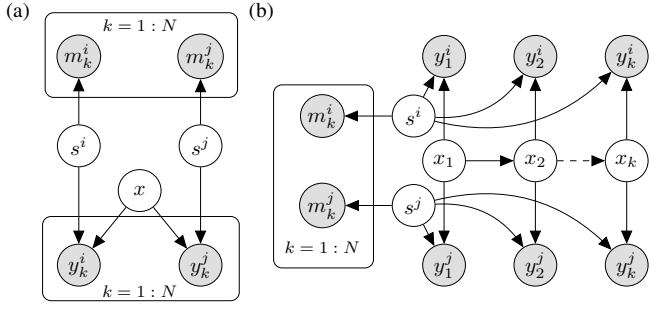
\begin{figure}[bt]
\resizebox{3.4in}{1.6in}{%
\begin{tikzpicture}[ new set=import nodes]
 \begin{scope}[nodes={set=import nodes}]
      \node (a) at (-2,3.25) {(a)};
      \node (s1)[latent] at (-1,1) {$s^i$};
      \node (s2)[latent] at (1,1) {$s^j$};
      \node (x)[latent] at (0,0.25) {$x$};
      \node (yi)[obs] at (-1,-1) {$y^i_{k}$};
      \node (yj)[obs] at (1,-1) {$y^j_{k}$};
      \node (mi)[obs] at (-1,2.5) {$m^i_{k}$};
      \node (mj)[obs] at (1,2.5) {$m^j_{k}$};
      { \tikzset{plate caption/.append style={below=20pt of #1.north }}
      \plate [inner sep=0.25cm, xshift=0.02cm, yshift=0.2cm] {plate1} {(yi) (yj)}       {$k=1:N$};}
      { \tikzset{plate caption/.append style={above=1pt of #1.north }}
      \plate [inner sep=0.2cm, xshift=0.02cm, yshift=-0.1cm, label={[shift={(-1.0,0)}]}] {plate2} {(mi)(mj)} {$k=1:N$};}

      \node (b) at (2,3.25) {(b)};
      \node (s3)[latent] at (4.5,1.5) {$s^i$};
      \node (s4)[latent] at (4.5,0) {$s^j$};
      \node (x1)[latent] at (5,0.75) {$x_1$};
      \node (x2)[latent] at (6.5,0.75) {$x_2$};
      \node (xk)[latent] at (8,0.75) {$x_k$};
      \node (yi1)[obs] at (5,2.5) {$y^i_{1}$};
      \node (yj1)[obs] at (5,-1) {$y^j_{1}$};
      \node (yi2)[obs] at (6.5,2.5) {$y^i_{2}$};
      \node (yj2)[obs] at (6.5,-1) {$y^j_{2}$};
      \node (yik)[obs] at (8,2.5) {$y^i_{k}$};
      \node (yjk)[obs] at (8,-1) {$y^j_{k}$};
      \node (mik)[obs] at (3,1.5) {$m^i_{k}$};
      \node (mjk)[obs] at (3,0) {$m^j_{k}$};
      
       { \tikzset{plate caption/.append style={below=70pt of #1.north }}
      \plate [inner sep=0.2cm, xshift=0.02cm, yshift=-0.05cm, label={[shift={(-1.0,0)}]}] {plate2} {(mik)(mjk)} {$k=1:N$};}
     
  \end{scope}
  
 \graph {
    (import nodes);
    {x,s1}->yi, {x,s2}->yj, s1->mi, s2->mj,
    
    % dynamic
    x1->x2, x2->[dashed]xk,
    {x1,s3}->yi1, {x1,s4}->yj1, s3->mik, s4->mjk,
    {x2}->yi2, {x2}->yj2, {xk}->yik, {xk}->yjk,
    s4->[bend left]yjk, s3->[bend right]yik, s3->[bend right]yi2,
    s4->[bend left]yj2
    };
    
\end{tikzpicture}}
\caption{(a) Static and (b) Partially dynamic Bayesian networks for two local random vectors $s^i, s^j$ (local measurement biases) and one common random vector $x$ (target state). In (a), $s^i$ and $s^j$ are conditionally independent given the static state $x$; in (b) the full time history $x_{1:k}$ is required for conditional independence.}
      \label{fig:GraphModel}
      \vspace{-0.2in}
\end{figure}

To motivate the problem, consider a simple static target tracking problem with one common target and two tracking agents as a running example (Fig. \ref{fig:GraphModel}a). 
Both agents $i$ and $j$ track the position of the common target, described by the random state vector $x$, and are assumed to have perfect knowledge of their own position, but unknown constant, non-zero, biases in the agent-target relative measurement position measurement described by the local random state vectors $s^i$ and $s^j$, respectively (similar to the bias in \cite{noack_treatment_2015}).

As shown in Fig. \ref{fig:GraphModel}a, at each time step $k$, the agents take two types of sensor measurements: 
(i) relative agent-target, described by $y^i_{k}$ and $y^j_{k}$ and (ii) $m^i_{k}$ and $m^j_{k}$ to different landmarks to locally collect data on their biases $s^i$ and $s^j$, respectively. It can be seen that the agents' local biases $s^i$ and $s^j$ become coupled due to measurements $y^i_{k}$ and $y^j_{k}$ of the common target $x$. Note that while we choose to use biases in the above example, the local, non-mutual, random states can be any other type of state, e.g., another target not mutually monitored by both agents.

In the case of homogeneous data fusion, the two agents preform inference over and communicate the full joint pdfs describing all rvs $p(x, s^i, s^j)$, including each other's local biases.
But in the heterogeneous fusion case, agents might hold a pdf over only a subset of the rvs $p^i(x,s^i)$, over the common rv (target $x$) and the agent's non-mutual rv (local bias $s^i$), making the dependencies between the non-mutual rvs \emph{hidden}. Therefore an agent might not be aware of the existence of another local bias random state vector (e.g., $s^j$) held by the other agent.
These dependencies are key to the problem, and the main challenge in heterogeneous fusion compared to homogeneous fusion is to account for them during fusion, where they are not explicitly represented in the local posterior pdfs.

\begin{comment}

\begin{figure}[tb]
      \centering
      %\framebox{\parbox{3in}{}
      \includegraphics[scale=0.42]{Figures/GraphModel2_v2.pdf}
      \caption{(a) Static and (b) Partially dynamic Bayesian networks for two local random vectors $s_i, s_j$ (local measurement biases) and one common random vector $x$ (target state). In (a), $s_i$ and $s_j$ are conditionally independent given the static state $x$; in (b) the full time history $x_{1:k}$ is required for conditional independence.}
      \label{fig:GraphModel}
      \vspace{-0.2in}
\end{figure}
\end{comment}
%\begin{comment}

%\end{comment}

\subsection{Problem Statement}
Assume a network of $n_a$ autonomous agents, performing recursive decentralized Bayesian updates to their prior pdf, with the goal of inferring the states of some global set of rvs $\chi_k\in \mathbb{R}^N$ at time $k$. Each agent $i\in \{1,...,n_a\}$ is tasked with the inference of a local subset of rvs $\chi^i_{k}\subseteq \chi_k$, which can be represented by an $n_i$-dimensional vector of random variables. An agent can update its local prior pdf for $\chi^i_{k}$, by (i) using Bayes' rule to recursively update a posterior pdf for $\chi^i_{k}$ with local sensor data $Y^i_{k}$ (e.g., $Y^i_{k}=[(y^i_{k})^T,(m^i_{k})^T]^T$ in Fig. \ref{fig:GraphModel}(a)) described by the conditional likelihood $p(Y^i_{k}|\chi^i_{k})$, and (ii) performing peer-to-peer fusion of external data $Z_{k}^{j,-}$ relevant to $\chi^i_{k}$ from any neighboring agent $j\in N_a^i$, where $Z_{k}^{j,-}$ is the local data agent $j$ has at time step $k$, prior to fusion with agent $i$ (i.e. from local sensor data $Y^i_{1:k}$ and from information received via fusion with other neighboring agents up to time $k-1$), and $N_a^i$ is the set of agents communicating with $i$.

The heterogeneous fusion problem now seeks a peer-to-peer fusion rule $\mathbb{F}$ which, given the local prior distribution $p^i(\chi^i_{k}|Z^{i,-}_{k})$ and a distribution $p^j(\chi^{ji}_{r,k}|Z^{j,-}_{k})$ over a subset of `relevant' random states from a neighboring agent $j$, returns a local fused conservative posterior distribution,
\begin{equation}
\begin{split}
        p^i_f(\chi^i_{k}|Z^{i,+}_{k}) = \mathbb{F}\big (p^i(\chi^i_{k}|Z^{i,-}_{k}), p^j(\chi^{ji}_{r,k}|Z^{j,-}_{k})\big ),
        \end{split}
    \label{eq:probStatement}
\end{equation} 
where $Z_{k}^{i,+}\equiv Z_{k}^{i,-}\cup Z_{k}^{j,-}$ is the combined data after fusion and $\chi^{ji}_{r,k}$ is the subset of random states at agent $j$ for which it has data to contribute to agent $i$, i.e., are relevant to agent $i$, and is assumed to be a non-empty set.
For instance, in the target tracking example, if $\chi^i_{k}=[x^T, (s^i)^T]^T$ and $\chi^j_{k}=[x^T, (s^j)^T]^T$, then the `relevant' random states in $j$ are $\chi^{ji}_{r,k}=x$. However, if $\chi^i_{k}=\chi^j_{k}=[x^T, (s^i)^T, (s^j)^T]^T$, but the agents collect data from local measurements (as in Fig. \ref{fig:GraphModel}(a)), agent $j$ might only have relevant data regarding the common target $x$ and its local bias $s^j$, thus $\chi^{ji}_{r,k}=[x^T, (s^j)^T]^T$. 

Both of the above examples can be represented by the heterogeneous fusion rule (\ref{eq:probStatement}). Thus \emph{heterogeneous fusion} encompasses the set of problems where the set of relevant rvs $\chi^{ji}_{r,k}$ is a subset of either agent $i$'s rvs $\chi^{ji}_{r,k}\subset \chi^i_{i,k}$, agent $j$'s rvs $\chi^{ji}_{r,k}\subset \chi^j_{k}$ or both. As a side note, the case where $\chi^i_{k}=\chi^j_{k}=\chi^{ji}_{r,k}$ (\ref{eq:probStatement}) simplifies to homogeneous fusion.

It is not immediately obvious how to to asses the validity of the sought fusion rule (\ref{eq:probStatement}), especially for heterogeneous fusion. Briefly, we look for a fusion rule which provides a \emph{conservative} posterior pdf, i.e., does not underestimate the uncertainty relative to the true pdf. In Gaussian pdfs for example, this means that the difference between the estimated covariance at agent $i$, $\bar{\Sigma}^i$ and a centralized estimator (having all the data from all agents), $\bar{\Sigma}^{cent}$, is positive semi-definite (PSD), i.e., $\bar{\Sigma}^i-\bar{\Sigma}^{cent}\succeq 0$. The reader is referred to Appendix \ref{sec:consist} for a more detailed discussion on the criteria for a `good fusion rule'.   

\subsection{Related Work}
\label{sec:realatedWork}
The idea of splitting a global state vector into $N$ subsets of $N_a$ vectors ($N_a \ll N$) has been posed to solve two different problems.
The first problem, as presented for example in \cite{chen_fusion_2002, noack_fusion_2014, klemets_hierarchical_2018, radtke_distributed_2019}, tries to reconstruct an MMSE \emph{point estimate} of the global state vector, by fusing $n_a$ local heterogeneous \emph{point estimates}. 
The second problem, which is the focus of this paper, tries to find $n_a$ different local \emph{posterior pdfs}, given all the information locally available up to the current time.

The problem of locally fusing data from a non-equal state vectors dates back to \cite{berg_model_1991}.
While this work offers a simple way for distributing into heterogeneous local state vectors, it assumes that the local pdfs are decoupled. 
In \cite{khan_distributed_2007}, Khan \etal relax that assumption, but ignore dependencies between states in the fusion step and restrict the state distribution to only states that are directly sensed by local sensors, where in practice, an agent might be `interested' in a larger set of local states that are not all directly sensed.
Similarly in cooperative localization \cite{li_cooperative_2012}, \cite{zhu_cooperative_2019} it is often assumed that agents are neighbors only if they take relative measurement to each-other. 
Then, the local state is augmented with the other agent's position states to process the measurement, often by assuming or approximating marginal independence. 
It can be shown that this approach represents a subset of the heterogeneous fusion problems considered in the paper.   

In \cite{chong_graphical_2004}, Chong and Mori use conditional independence in Bayesian networks to reduce the subset of states to be communicated. 
However, they assume hierarchical fusion and for dynamic systems only consider the case of deterministic state processes, omitting the important case of stochastic processes which is considered in this paper.
The work by Whitacre \etal \cite{whitacre_decentralized_2011}, does not formally discuss conditional independence but introduces the idea of marginalizing over common states to fuse Gaussian distributions, when cross correlations of common states are known. 
However, they implicitly assume cross correlations among the conditioned, or unique, states are small. 
While this assumption might hold for their application, it does not offer a robust solution. 
Reference \cite{ahmed_factorized_2016} uses CI with Gaussian distributions to relax the assumption of known correlations of the fused state. 
Reference \cite{saini_decentralized_2019} suggests a similar solution to \cite{ahmed_factorized_2016} but restrict it to systems where all agents have the same set of common states. Although scalable and simple, such approaches do not account for dependencies between locally exclusive states. 

\begin{comment}
These dependencies are key to the problem, since they lead to coupled uncertainties and correlated estimation errors. For example, consider a simple target tracking problem with one target and two tracking agents. Assume the agents have perfect ownship position information, but unknown constant biases $s_i$ and $s_j$ in the agent-target relative measurement vectors $y_{i,k}$ and $y_{j,k}$, respectively. The agents can also take relative observations $m_{i,k}$ and $m_{j,k}$ to different landmarks to locally estimate these ownship measurement biases. As shown in Fig. \ref{fig:GraphModel}, two agents $i$ and $j$ become coupled due to the common target measurements. In the homogeneous case, the two agents process and communicate each other's ownship bias states in addition to the target states and explicitly account for their dependencies. But in the heterogeneous case, each agent is assumed to only estimate target states and its local own biases. The question is how to account for these dependencies during fusion, when these are not explicitly present in the local posterior pdfs.
\end{comment}

This work aims at gaining insight and understanding on how such issues should be addressed by exploiting the structure of the underlying dynamical system in such problems. 
\section{Heterogeneous Fusion}
\label{sec:factorCF} 
\begin{figure*}[tb!]
	\centering
	\includegraphics[width=0.9\textwidth]{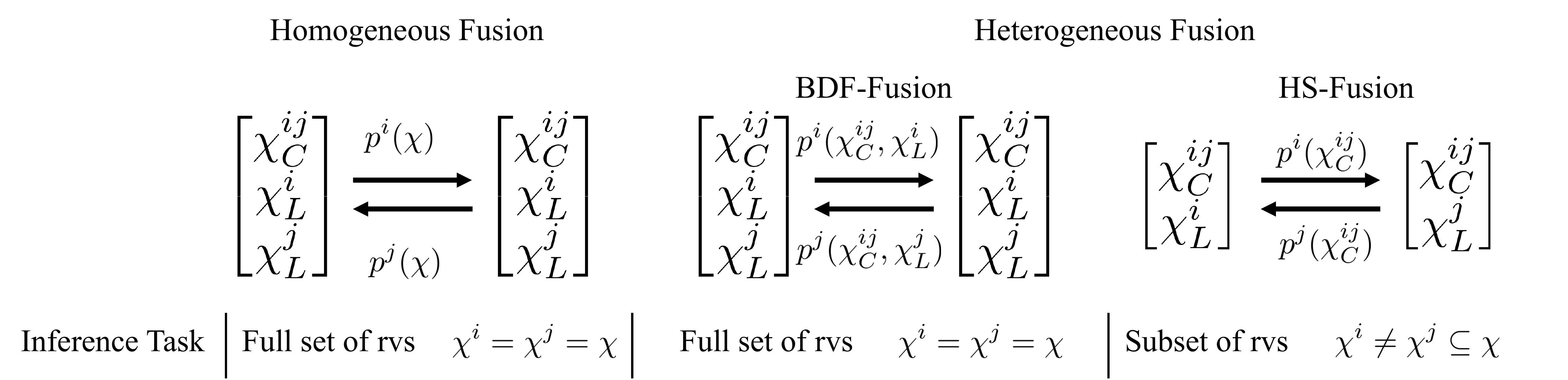}
	\caption{Progression of fusion rules derived in this paper, describing the set of rvs which needs to be: (i) communicated between any two agents in an undirected acyclic graph; (ii) inferred (estimated) by each agent, corresponding to computation load.}
	%\caption{Different heterogeneous fusion rules derived in this paper. Starting from homogeneous fusion over the full set of rvs, the Bi-directional-factorized (BDF) fusion rule is developed, where agents infer the full set of rvs, but communicate pdfs over unequal subsets. Then agents are tasked with subset of rvs and communicate pdfs over the common subset in the heterogeneous-state (HS) fusion rule, reducing computation and communication load significantly. }
	\label{fig:fusionChart}
	\vspace{-0.2in}
\end{figure*}

%Many problems in robotics can be viewed as an inference problem, where the unknown variables are described by a probability density function (pdf) \cite{attias_planning_2003}, \cite{dellaert_factor_2021}. While these pdfs are often approximated as Gaussian pdfs, the true distribution might might be non-Gaussian (e.g., \cite{ong_decentralised_2008}). To develop heterogeneous fusion rules that 
The analysis starts with a general probabilistic formulation of the homogeneous fusion rule. Then as seen in Fig. \ref{fig:fusionChart}, different fusion problems of interest are discussed. Starting from the homogeneous fusion rule, where all agents keep and share a posterior pdf of the full global joint random state vector $\chi$, cases are then considered where an agent is only interested in and/or observes a subset of the full set of rvs. The key idea is to exploit the structure of the problem, specifically, conditional independence between local sets of rvs. Assuming the dependencies in the data are explicitly tracked, an exact heterogeneous fusion rule, the \emph{Bi-directional factorized fusion} (BDF-fusion) is developed, in which an agent still infers the global set of rvs, but needs to only communicate a subset of them.  
By discarding dependencies to non-local rvs, at the cost of approximation of the full global joint pdf, agents are able to hold pdfs over overlapping subsets of rvs. This \emph{Heterogeneous state fusion} (HS-fusion) rule further improves the scalability of the system, as agents only infer a subset of rvs and communicate marginal pdfs over common variables.   
Thus, it scales with the size of the local subset of rvs an agent monitors and not with the size of the system's global set of rvs, which leads to significant communication and computation savings.

\subsection{Homogeneous Decentralized Fusion}

\begin{figure}[tb]
      \centering
      %\framebox{\parbox{3in}{}
      \includegraphics[scale=0.55]{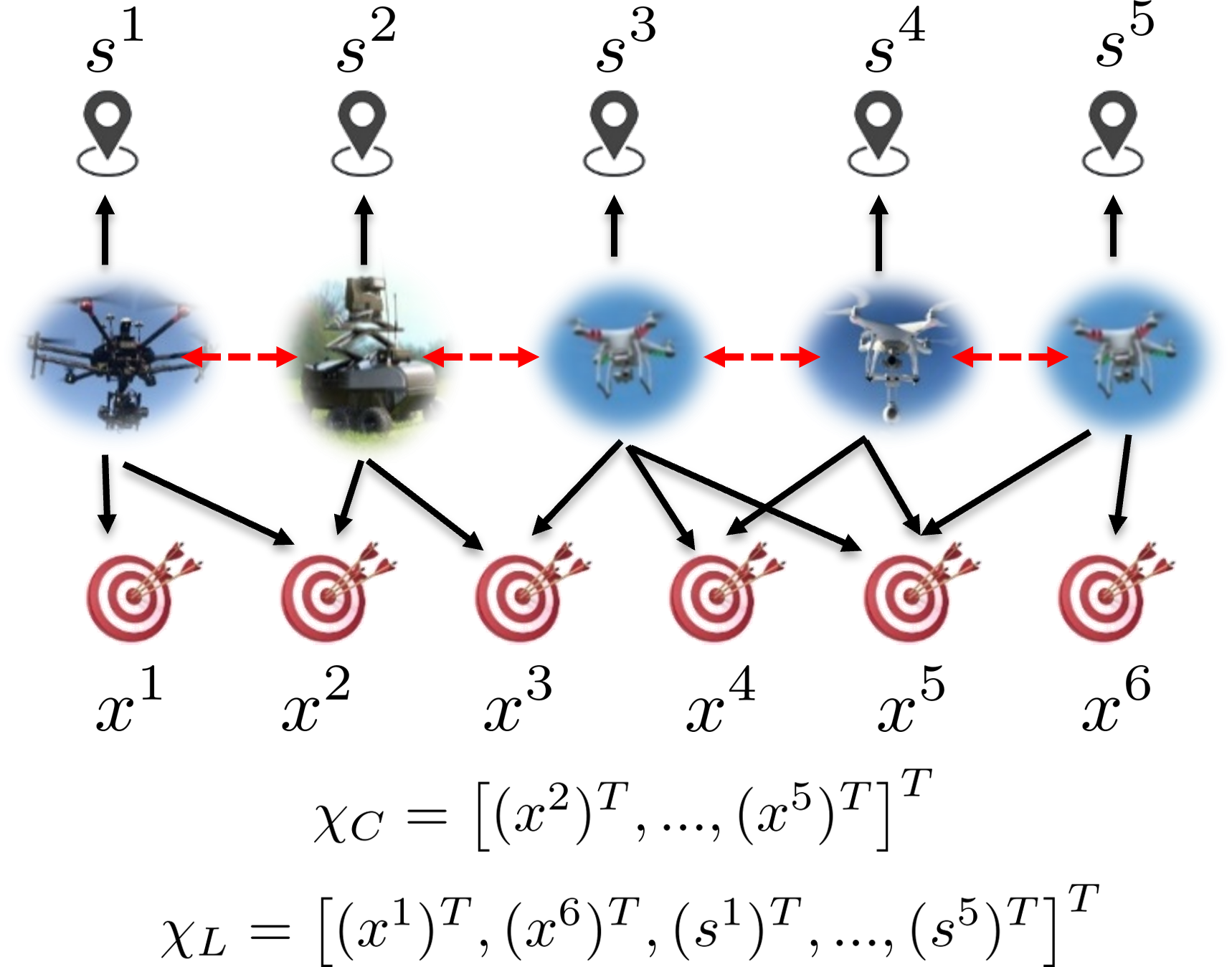}
      \caption{Target tracking example. Full black arrows denote local measurements to targets $x^t$ and landmarks to estimate local biases $s^i$, red dashed arrows indicate bi-directional communication channel between agents. The undirected and a-cyclic chain topology can be seen.}
      \label{fig:targetTrackingExample}
      \vspace{-0.2in}
\end{figure}

Let the full global set of rvs be $\chi=\chi_C\cup \chi_L$ describe a set of random states described by the joint pdf, $p(\chi)=p(\chi_C, \chi_L)$, where $\chi_C$ and $\chi_L$ are all the common and local states in the system, respectively, represented by state vectors. For example, in the target tracking example in Fig. \ref{fig:targetTrackingExample}, $\chi_C=[(x^2)^T,...,(x^{5})^T]^T$ and $\chi_L=[(x^1)^T,(x^6)^T,(s^1)^T,...,(s^{5})^T]^T$.
For now it is assumed that the random states, thus $\chi$, are static and the time index notation is dropped (dynamic states will be revisited and considered later).

\begin{comment}

\begin{equation}
    \chi=\begin{bmatrix}X\\ S\\
    \end{bmatrix} \ \ \ \in\mathbb{R}^{(n_X+n_S)\times1},
    \label{eq:stateDef}
\end{equation}

\end{comment}

The goal is to find the fused estimated underlying distribution of the random state vector $\chi$ given the joint data $Z^+_k=Z^{i,-}_{k}\cup Z^{j,-}_{k}$ at agents $i$ and $j$. Using a distributed variant of Bayes' rule, \cite{chong_distributed_1983} shows that the exact posterior pdf conditioned on the joint data of $i$ and $j$, i.e., the homogeneous peer-to-peer fusion rule, is given by:
\begin{equation}
    p_f(\chi|Z_k^+)\propto \frac{p^i(\chi|Z^{i,-}_{k})p^j(\chi|Z^{j,-}_{k})}{p^{ij}_c(\chi|Z_{k}^{i,-}\cap Z_{k}^{j,-})},
    \label{eq:bayesCF}
\end{equation}
where $p^{ij}_c$ is the posterior pdf conditioned on the common data shared by agents $i$ and $j$ prior to the current fusion.

The key challenge in computing the posterior fused pdf in (\ref{eq:bayesCF}) is to correctly account for the pdf in the denominator, $p^{ij}_c(\chi|Z_{k}^{i,-}\cap Z_{k}^{j,-})$. There are several method to approximate it, e.g., CI \cite{julier_non-divergent_1997}, GMD \cite{bailey_conservative_2012}, when dependencies in the data held by the two agents is unknown. However, this work seeks to explicitly track those dependencies, thus uses the CF to recursively compute $p^{ij}_c(\chi|Z_{k}^{i,-}\cap Z_{k}^{j,-})$.

In the following, the structure of the underlying estimation problem is exploited to conditionally factorize into relevant subsets of the global random state vector, thereby enabling reduced communication costs and allowing each agent to locally hold a smaller pdf, i.e., reducing the computational cost of inference. 
\subsection{ Exact Heterogeneous Fusion}
\begin{comment}

\odb{From the law of total probability, the joint pdf over $\chi$ can be conditionally factorized as
\begin{equation}
    p(\chi)=p(\chi_C,\chi_L)=p(\chi_C)\cdot p(\chi_L|\chi_C).
    \label{eq:factorDist}
\end{equation}
Using this factorization, (\ref{eq:bayesCF}) can be expressed as
\begin{equation}
    \begin{split}
     p_f(\chi|Z_k^+)\propto 
     &\frac{p^i(\chi_C|Z_{k}^{i,-})p^j(\chi_C|Z_{k}^{j,-})}{p_c(\chi_C|Z_{k}^{i,-}\cap Z_{k}^{j,-})}\\
     &\cdot\frac{p^i(\chi_L|\chi_C,Z_{k}^{i,-})p^j(\chi_L|\chi_C,Z_{k}^{j,-})}{p_c(\chi_L|\chi_C,Z_{k}^{i,-}\cap Z_{k}^{j,-})}.
     \end{split}
     \label{eq:exactFusion}
\end{equation}
Here the peer-to-peer fusion rule (\ref{eq:bayesCF}) is simply separated into the product of common states fusion and conditional local states fusion. }
\end{comment}

To simplify the derivation, let us consider for now a two-agent problem, with $\chi_L=\chi_L^i\cup \chi_L^j$, where for example $\chi^i_L=s^i$, $\chi^j_L=s^j$ and $\chi_C=\chi_C^{ij}=x$ (as in Fig. \ref{fig:GraphModel}(a)).
The details for more than 2 agents are given in Sec. \ref{sec:algo}.
From the law of total probability, the joint pdf over $\chi$ can be conditionally factorized as
\begin{equation}
    p(\chi)=p(\chi_C^{ij},\chi_L^i,\chi_L^j)=p(\chi_C^{ij})\cdot p(\chi_L^i|\chi_C^{ij})\cdot p(\chi_L^j|\chi_C^{ij},\chi_L^i).
    \label{eq:factorDist}
\end{equation}
As in many robotic applications, some tasks are local in nature \cite{dellaert_factor_2021}, i.e. agents locally gather data regarding those tasks. Thus, agent $i$ might collect data regarding agent $j$'s local states $\chi^j_L$ only via communication with agent $j$ and vice versa. Further, as can be seen from the probabilistic graphical model (PGM) in Fig. \ref{fig:GraphModel}(a), given the common state $\chi_C^{ij}=x$, the local states and data collected by agent $i$ ($\chi_L^i=s^i$ and $Z_{k}^{i,-}$) are conditionally independent from the local states and data at agent $j$ ($\chi_L^j=s^j$) and $Z_{k}^{j,-}$, 
\begin{equation}
    \chi_L^i,Z_{k}^{i,-}\perp \chi_L^j,Z_{k}^{j,-}|\chi_C^{ij}\Rightarrow     s^i,y^i_{k},m^i_{k}\perp s^j,y^j_{k},m^j_{k}|x.
    \label{eq:condInd}
\end{equation}
Therefore, the factorization in (\ref{eq:factorDist}) can be further simplified for agents $i$ and $j$,
\begin{equation}
    \begin{split}
    &p^i(\chi|Z^{i,-}_k)= p^i(\chi_C^{ij}|Z^{i,-}_k)\cdot p^i(\chi_L^i|\chi_C^{ij},Z^{i,-}_k)\cdot p^i(\chi_L^j|\chi_C^{ij}) \\
    &p^j(\chi|Z^{j,-}_k)=p^j(\chi_C^{ij}|Z^{j,-}_k)\cdot p^j(\chi_L^i|\chi_C^{ij})\cdot p^j(\chi_L^j|\chi_C^{ij},Z^{j,-}_k).
    \end{split}
    \label{ij_factorization}
\end{equation}
The homogeneous fusion rule in (\ref{eq:bayesCF}) can then be conditionally factorized as,
\begin{equation}
    \begin{split}
     &p_f(\chi|Z^+_k)\propto \overbrace{ \frac{p^i(\chi_C^{ij}|Z_{k}^{i,-})p^j(\chi_C^{ij}|Z_{k}^{j,-})}{p_c^{ij}(\chi_C^{ij}|Z_{k}^{i,-}\cap Z_{k}^{j,-})}}^{p_f(\chi_C^{ij}|Z^{+}_k)}\cdot \\  &\underbrace{\frac{p^i(\chi_L^i|\chi_C^{ij},Z_{k}^{i,-})p^j(\chi_L^i|\chi_C^{ij})}{p_c^{ij}(\chi_L^i|\chi_C^{ij})}}_{p_f(\chi_L^i|\chi_C^{ij},Z^{i,-}_k))}\cdot \underbrace{  \frac{p^i(\chi_L^j|\chi_C^{ij})p^j(\chi_L^j|\chi_C^{ij},Z_{k}^{j,-})}{p_c^{ij}(\chi_L^j|\chi_C^{ij})}}_{p_f(\chi_L^j|\chi_C^{ij},Z^{j,-}_k)},
    \end{split}
    \label{eq:exactFusion_3states}
\end{equation}
where to simplify the pdf in the denominator we used the fact that if $\chi^i_L$ is conditionally independent from $Z_{k}^{j,-}$, it is also conditionally independent from the intersection of $Z_{k}^{j,-}$ and $Z_{k}^{i,-}$, i.e. $\chi_L^i\perp Z_{k}^{i,-}\cap Z_{k}^{j,-} |\chi_C^{ij}$.

The expressions for the local states fusion $p_f(\chi_L^i|\chi_C^{ij},Z^{i,-}_k)$ and $p_f(\chi_L^j|\chi_C^{ij},Z^{j,-}_k)$ can be further simplified by recognizing: 
\begin{equation*}
    p^j(\chi_L^i|\chi_C^{ij}) = p_c^{ij}(\chi_L^i|\chi_C^{ij}) \ \mathrm{and} \ p^i(\chi_L^j|\chi_C^{ij}) = p_c^{ij}(\chi_L^j|\chi_C^{ij}),
\end{equation*}
which simply states that if the pdf in the denominator is explicitly tracked, it should equal the conditional pdf over the local rv held by the agent not monitoring that rv.

The heterogeneous \emph{BDF-fusion} rule can then be written as,
\begin{equation}
    \begin{split}
     p_f(\chi|Z^+_k)\propto  p_f(\chi_C^{ij}|Z^{+}_k)p^i(\chi_L^i|\chi_C^{ij},Z_{k}^{i,-}) p^j(\chi_L^j|\chi_C^{ij},Z_{k}^{j,-}),
    \end{split}
    \label{eq:BDF_fusion}
\end{equation}
where $p_f(\chi_C^{ij}|Z^{+}_k)$ is defined in (\ref{eq:exactFusion_3states}).
The BDF-fusion equation agrees with the intuition described before: if, when conditioning on the common states $\chi_C^{ij}$, agent $j$ ($i$) does not gain any new local data regarding agent $i$'s ($j$'s) local state $\chi_L^i$ ($\chi_L^j$), it need not communicate its respective local conditional pdf $p^j(\chi^i_L|\chi^{ij}_C)$ ($p^i(\chi^j_L|\chi^{ij}_C)$). 

The transition from the homogeneous fusion rule to the heterogeneous BDF-fusion rule is shown in Fig. \ref{fig:fusionChart}, and is enabled by using the underlying conditional independence structure of the problem to separate common and local variables.
This fusion rule provides a possible $\mathbb{F}$ sought in (\ref{eq:probStatement}): here, the communicated distributions are over different sets of random states, as agent $i$ sends the marginal pdf $p^i(\chi^{ij}_C,\chi_L^i|Z_k^{i,-})$ and receives from $j$ the marginal $p^j(\chi^{ij}_C,\chi_L^j|Z_k^{j,-})$. The sets of relevant states are then $\chi^{ij}_{r,k}=\chi^{ij}_C\cup \chi_L^j$ and $\chi^{ji}_{r,k}=\chi^{ij}_C \cup \chi_L^j$ when fusing at agent $i$ and $j$, respectively.

Note that while the fused posterior pdf in (\ref{eq:BDF_fusion}) is equivalent to the one achieved by the homogeneous fusion rule (\ref{eq:bayesCF}) (for static systems, or when the full time history is maintained), there are two key distinctions between them. First, the math behind this heterogeneous fusion rule is fundamentally different.
In the latter, the conditional prior pdf regarding the non-local states $\chi_L^j$ at agent $i$ (and similarly at $j$) is simply replaced by the pdf received from $j$, $p^j(\chi_L^j|\chi^{ij}_C,Z_k^{j,-})$, which treats the common random state $\chi^{ij}_C$ as a function parameter. 
Then, the \emph{fused} marginal is recombined with the two conditional pdfs in the joint distribution via the law of total probability (\ref{eq:factorDist}) at each agent. 
This weights the conditional pdf differently as a function of $\chi^{ij}_C$ and changes the overall joint pdf, thus implicitly/indirectly updating the \emph{marginal} pdf over $\chi_L^i$. For example, in the target tracking example, that means that while agent $i$ does not directly receive new data from $j$ regarding its local bias $s^i$, its marginal pdf over $s^i$ gets updated due to the fusion of data over $x$ and $s^j$. Second, the BDF-fusion rule offers considerable reduction in communication requirements achieved by sending only new and relevant data (given that the above assumptions are met), as shown later in Sec. \ref{sec:NumExample}.

\subsection{Approximate Heterogeneous Fusion}

So far we considered a set of problems where all agents across the network hold a local pdf over the same global random state vector $\chi$, which includes all locally relevant random states. This requires each agent to hold a local pdf over the full global random state vector and to communicate the conditional pdfs regarding its local random states. Despite its potential communication reduction relative to the homogeneous fusion rule, it might still require considerable communication volume, for example, when $\chi^i_L$ are agent $i$'s local states for a navigation filter, which typically has 16 or more states \cite{dourmashkin_gps-limited_2018}. By allowing each agent to only reason about a subset of states, e.g., only states within its inference task, significant reductions in both computation and communication requirements can be gained. This motivates the development of an approximate heterogeneous fusion rule that scales with the agent tasks and not the global network tasks (or number of agents), we dub it \emph{Heterogeneous state} (HS) fusion.

% , thus, as the number of tasks in the network increases, so does the local computation load
More formally, the set of problems where each agent only holds a pdf over its locally relevant subsets of states $\chi^i\subset \chi$ is considered. Here the heterogeneous subsets of random states are defined as $\chi^i=\chi^{ij}_C\cup\chi^i_L$ and $\chi^j=\chi^{ij}_C\cup\chi^j_L$ for agents $i$ and $j$, respectively. By marginalizing out the set of `irrelevant' states, $\chi_L^j$ ($\chi_L^i$), the fusion rule for each agent, over their locally relevant random states, can be written as 
\begin{equation}
    \begin{split}
     &p^i_{f}(\chi^i|Z_k^{i,+})\propto p_f(\chi_C^{ij}|Z^{+}_k)\cdot p^i(\chi_L^i|\chi_C^{ij},Z_{k}^{i,-})\\
     &p^j_{f}(\chi^j|Z_k^{j,+})\propto p_f(\chi_C^{ij}|Z^{+}_k)\cdot p^j(\chi_L^j|\chi_C^{ij},Z_{k}^{j,-}).
    \end{split}
    \label{eq:HeteroFusion_3states_i}
\end{equation}

HS-fusion gives another fusion rule $\mathbb{F}$ for the problem statement in (\ref{eq:probStatement}), where the sets or relevant states are now $\chi^{ij}_{r,k}=\chi^{ji}_{r,k}=\chi^{ij}_C$. Notice that while the two pdfs are over different sets of random states, the marginal pdfs over the common random state vector $\chi_C^{ij}$, held by both agents will be equal, i.e, $p_{f}^i(\chi_C^{ij}|Z_k^{i,+}) = p_{f}^j(\chi_C^{ij}|Z_k^{j,+})$. The advantages of this fusion rule in scalability is demonstrated in Sec. \ref{sec:NumExample}.

\subsection{Fusion Algorithm}
\label{sec:algo} 

Decentralized fusion algorithms, in general, are built out of two main steps: sending out a message and fusion of incoming messages. 
In homogeneous fusion rules, messages are over the same full state vector $\chi$. 
In heterogeneous fusion, on the other hand, either the communicated or the local distributions are over different random state vectors. Thus there is a need to clarify what are the step that an agent $i$ needs to perform to locally construct and fuse messages to or from its $N_a^i\geq2$ neighboring agents.

\subsubsection{\textbf{Constructing Messages}}
In BDF-fusion, an agent \emph{i} holds a posterior distribution over the full global random state vector $\chi$. 
Assuming the communication topology for the network of agents is an acyclic undirected graph, agent $i$ needs to communicate to any of its neighboring agents $j\in N^i_a$ a distribution over the set of local states $\chi^i=\chi^i_L\cup\chi_C^{ij}$ and the set $\chi_{\neg i}^{ij}$ (see Sec.\ref{sec:passThrough} for definition and explanation). On the other hand, in HS-fusion, agent \emph{i}'s local distribution is only over the set of local relevant states $\chi^i$, and only communicates the common subset $\chi_C^{ij}$. 
Agent \emph{i} thus sends agent \emph{j} the following marginal distributions,
\begin{equation}
    \begin{split}
        &\text{BDF-CF:} \ \ p^i_{ij}(\chi^i\cup\chi_{\neg i}^{ij})=\int p^i(\chi)d\chi_{\neg i} \ \ \ \ \ \ \ \ \  \forall j \in N^i_a	\\
        &\text{HS-CF:} \ \ p^i_{ij}(\chi_{C}^{ij})=\int p^i(\chi^i)d(\chi^i\setminus \chi_C^{ij})\ \ \ \ \ \ \ \ \  \forall j \in N^i_a,
    \end{split}
    \label{eq:message}
\end{equation}
where $\chi_{\neg i}=\chi\setminus \{\chi^i\cup\chi_{\neg i}^{ij}\}$ and
the dependency on the data $Z^i$ is implied from here on for brevity and will be explicitly shown if needed.

\subsubsection{\textbf{Fusing Messages}}
Since, in general, agent $i$ has a different sets of random states in common, $\chi_C^{ij}$, with any of its neighboring agents $j\in N_a^i$, the local fusion equations requires multiplying (and dividing) pdfs over different sets of random states. 
The following equations detail the heterogeneous fusion operation from the perspective of an agent $i$, communicating with its $n_a^i$ neighboring agents, %,\footnote{While the equations are written in logarithmic form for the aforementioned reasons, they can be easily implemented using conventional pdf form. }
\begin{comment}

\begin{equation}
    \begin{split}
    \text{BDF-CF:} \ \ \ \\
    \log[p^i_{f}(\chi)]& = \log[p^i(\chi^i\cup\chi_{\neg i}^{ij})]+\\
    &\sum_{j\in N^i_a}\log[p^j_{ji}(\chi^j\cup\chi_{\neg  j}^{ji})]-\log[p_c^{ji}(\chi_{C}^{ji})],\\
    \text{HS-CF:} \ \ \ \ \\
    \log[p^i_{f}(\chi^i&)] = \log[p^i(\chi^i)]+\\
    &\sum_{j\in N^i_a}\log[p^j_{ji}(\chi_{C}^{ji})]-\log[p_c^{ji}(\chi_{C}^{ji})].
    \end{split}
    \label{eq:setFusion}
\end{equation}
\end{comment}

\begin{equation}
    \begin{split}
    \text{BDF-CF:} \ \ \ \\
    p^i_{f}(\chi)& = p^i(\chi^i\cup\chi_{\neg i}^{ij})\cdot
    \prod_{j\in N^i_a}\frac{p^j_{ji}(\chi^j\cup\chi_{\neg  j}^{ji})}{p_c^{ji}(\chi_{C}^{ji})},\\
    \text{HS-CF:} \ \ \ \ \\
    p^i_{f}(\chi^i&) = p^i(\chi^i)\cdot
    \prod_{j\in N^i_a}\cdot\frac{p^j_{ji}(\chi_{C}^{ji})}{p_c^{ji}(\chi_{C}^{ji})}.
    \end{split}
    \label{eq:setFusionPDF}
\end{equation}

Note that for these equations to be valid, the local subsets must be conditionally independent given the common subsets (\ref{eq:condInd}), in dynamic systems, this presents a challenge. We further discuss and suggest a solution in Sec. \ref{sec:DynamicSystems}. 

\subsubsection{\textbf{Passed-Through States}}
\label{sec:passThrough}
\begin{figure}[tb]
	\centering
	\includegraphics[width=0.2\textwidth]{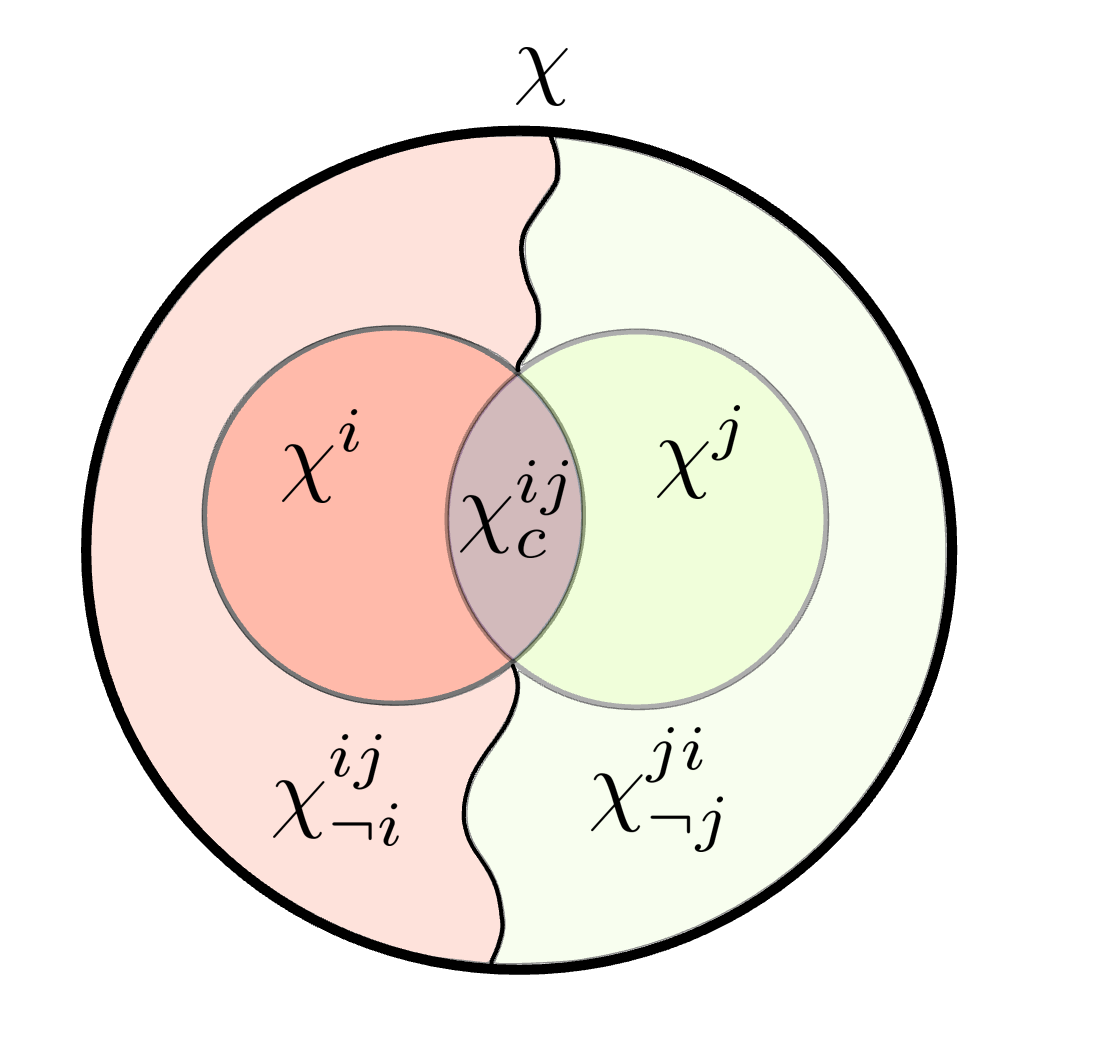}
	\caption{Diagram presenting the division of the full random state vector $\chi$ into smaller subsets.}
	\label{fig:setDiagram}
	\vspace{-0.2in}
\end{figure}

In heterogeneous fusion agents communicate and/or process data with respect to (w.r.t.) relevant tasks, where recall that relevant tasks are those for which they have new data to communicate to their neighbor. These data might be from local observation or observed by another agent down-stream in the network and `passed through' to the up-stream agent.
For example, in Fig. \ref{fig:targetTrackingExample}, when agent 2 communicates with agent 1, it gains data by local observations of targets 2,3 and its local bias and `passes through' data regarding targets 4,5,6 and biases 3,4,5. We define these random states as $\chi_{\neg i}^{ij}$ for states that are to be passed from \emph{i} to \emph{j} but are not local to \emph{i}, and similarly for $\chi_{\neg j}^{ji}$. 
With these definitions, as illustrated in Fig. \ref{fig:setDiagram}, the following holds,
\begin{equation*}
    \chi=\chi^i+\chi^j-\chi_c^{ij}+\chi_{\neg i}^{ij}+\chi_{\neg j}^{ji}.
\end{equation*}

\section{Conditional Independence in Dynamic Systems}
\label{sec:DynamicSystems}
To describe and analyze the problem of conditional independence for heterogeneous fusion in dynamic systems, we assume (without loss of generality) that the local subset $\chi_L$ describe static rvs, and that the common subsets $\chi_C$ describes dynamic rvs. Note that the opposite case, where the local subset is dynamic and common subset is static, is simpler, since filtering preserves the required conditional independence assumption.

For the heterogeneous fusion algorithms in (\ref{eq:setFusionPDF}) to be valid, the local subsets must be conditionally independent given the common subsets (\ref{eq:condInd}). However, in problems corresponding to dynamic or partially-dynamic Bayesian networks, e.g., in Fig. \ref{fig:GraphModel}(b), it is generally not possible to claim conditional independence of local states (or local data) based on the \emph{filtered} dynamic state. In Fig. \ref{fig:GraphModel}(b), $s_i\not\perp s_j|x_k$, i.e., $s_i$ and $s_j$ are not conditionally independent given $x_k$ when it is successively marginalized over time. Thus there is a need to regain conditional independence in dynamic stochastic systems for the above fusion rules to work correctly. 
%The problem of regaining conditional independence is first discussed for general pdfs, then a closed form solution is given for the case of linear Gaussian systems. The different steps each agent takes to perform heterogeneous fusion are summarized in a pseudo-code (Algorithm \ref{algo:CF2}) using the closed-form operations for Gaussian pdfs.}

There are two approaches to solve this problem: (i) keeping a distribution over the full time history  $p(\chi_{C,k:1},\chi_{L}|Z_k)$, where $\chi_{C,k:1}$ denotes all common dynamic rvs from $k=1$ until current time step $k$; (ii) enforcing conditional independence after marginalization by conservative approximations which disconnects the dependencies between the relevant random states. 
In the following, these solutions for the case of general pdfs are discussed. Then, Sec. \ref{subSec:Gaussians} derives specific %solutions and 
closed-form representations for Gaussian distributions, and a pseudo-code, summarizing the different steps each agent takes to perform heterogeneous fusion is given in Algorithm \ref{algo:CF2}.

\subsection{Augmented State}
The distribution over the full augmented random state $\chi_{k:1}=\chi_{C,k:1}\cup\chi_{L}$, given the data $Z_k^{i,-}$ can be recursively updated using the following formula \cite{koch_accumulated_2011}:
\begin{equation}
    p^i(\chi_{k:1}|Z_k^{i,-})\propto p(\chi_{k-1:1}|Z_{k-1}^{i,+})p(\chi_{C,k}|\chi_{C,k-1})p(Y_k^i|\chi_k),
    \label{eq:pdf_aug_state}
\end{equation}
where here $Z_{k-1}^{i,+}$ is used to indicate an agent's data after the previous fusion step, $k-1$, $Y_k^i$ for the local sensor data gained at the current time step $k$, and $Z_k^{i,-}$ is the data at time step $k$ prior to fusion. 

Keeping the full time history of the dynamic common states maintains the conditional independence assumption, as can be seen for example in Fig. \ref{fig:GraphModel}(b). Here $\chi_{C,k:1}=x_{k:1}$, $\chi_L=[(s^i)^T,(s^j)^T]^T$, and the conditional independence assumption $s^i\perp s^j|x_{k:1}$ holds.

While the augmented state approach leads to an increase in the communication and computation requirements as the size of the state vector $\chi_{k:1}$ increases,  efficient inference algorithms that exploit the Markovian property of the dynamic system and its structure can be designed. For example, for Gaussian distributions, the information matrix structure is close to block-diagonal (see Sec. \ref{subSec:iAS}), i.e., marginalization can be done efficiently. Furthermore, the communication requirement can be bounded, as in practice agents do not need to communicate the full state time history. Instead the communicated pdf can be reduced to be on a time window relative to the size of the network; this guarantees information can propagate to the `far' end of the network graph.
For example, in the chain network in Fig. \ref{fig:targetTrackingExample}, the data that agent 1 gathers at time step 1 will reach agent 6 after 5 communication steps. Thus after 5 time steps, there is no new data to communicate over the target position at time step 1.

\subsection{Conservative Filtering}
Full knowledge over past system states has been assumed thus far, which enables conditional independence between two agents local states and the derivation of new heterogeneous fusion rules. However, in many distributed fusion applications it is desirable to maintain only a limited time window of recent state history. 
Thus, marginalizing out past states into a smaller sliding window of recent time steps might be favored, as maintaining the full accumulated state densities results in rapid state dimension growth and yields computation and communication burden. 
While marginalization is rather trivial for homogeneous fusion problems, in heterogeneous fusion extra care must be taken to maintain conditional independence. 
Without loss of generality, for the rest of the paper, a small sliding window of only the current time step (as done in the Kalman-filter (KF) for example) will be used.

Given a joint distribution $p(\chi_{C,2:1},\chi_L^i,\chi_L^j)$, described by the PGM of Fig. \ref{fig:GraphModel}(b), for example, marginalizing out $\chi_{C,1}=x_1$, as done in filtering, results in coupling of all the variables in its Markov blanket, $\chi_{C,2}=x_2$, $\chi_L^i=s^i$ and $\chi_L^j=s^j$. 
Since conditional independence between $\chi_L^i$ and $\chi_L^j$ is an imperative assumption in the basis of the proposed fusion rules, it is necessary to retain it after marginalization. 
Thus, the goal is to approximate the \emph{dense} distribution $p(\chi_{C,2},\chi_L^i,\chi_L^j)$ by a conservative \emph{sparse} distribution,
\begin{equation}
    \begin{split}
    &\Tilde{p}(\chi_{C,2},\chi_L^i,\chi_L^j)\propto p(\chi_{C,2})p(\chi_L^i|\chi_{C,2})p(\chi_L^j|\chi_{C,2}).
    \end{split}
\end{equation}
For the pdf to be consistent, the approximation $\Tilde{p}(\chi_{C,2},\chi_L^i,\chi_L^j)$ has to be conservative w.r.t. $p(\chi_{C,2},\chi_L^i,\chi_L^j)$. Loosely speaking, this means the approximate distribution $\Tilde{p}(\cdot)$ does not underestimate the uncertainty of the true distribution $p(\cdot)$. 
A more detailed discussion and definitions of consistency and conservativeness as treated in this paper is given in Appendix \ref{sec:consist}. 

The above discussion is brought here for completeness, but the treatment of conservative sparse marginalization of general pdfs is out of scope of this paper and is left for future work. However, for Gaussian pdfs, where there is agreement of the definition of conservativeness, a conservative sparse marginalization solution is derived in the next section (Sec. \ref{subSec:GaussianSparseMarg}).

\section{Heterogeneous Channel Filter - A Closed Form Algorithm}
\label{subSec:Gaussians}

Thus far, the problem and derivation are stated in general pdf terms. From here, to further the understanding and increase the intuition of the heterogeneous fusion problem, we explicitly track the information flow and dependencies in heterogeneous DDF problems by: (i) extending the (homogeneous) channel filter (CF) algorithm to heterogeneous DDF; (ii) focus our attention to the information (canonical) form of the Gaussian distribution, and linear models.

\textbf{The homogeneous CF:} The CF is a simple method to track dependencies in the data for a network of agents performing \emph{homogeneous} DDF \cite{grime_data_1994}. Over the past two decades, the CF algorithm has been used in many practically fielded robotics applications, such as search tasks in multi-UAV \cite{bourgault_communication_2004} and human-robot teams \cite{bourgault_scalable_2008}, terrain estimation  \cite{schoenberg_distributed_2009}, and multi-target tracking \cite{ong_decentralised_2008} where it has been used to fuse information between particle filters.
The core idea of the CF is to add a filter on the communication channel between any pair of agents. This filter explicitly tracks $p^{ij}_c$ (\ref{eq:bayesCF}), the posterior pdf conditioned on the common data shared (over the channel) by agents $i$ and $j$. 
For \emph{homogeneous} fusion, \cite{grime_data_1994} shows that each agent is able to recover the optimal centralized state estimate if: (i) the communication graph between the $n_a$ agents is undirected and acyclic (such that data does not circle back to its sender), e.g., tree or chain communication topology; (ii) there is full rate communication and sequential processing of incoming data, i.e., an agent sends a message at each time step $k$ without delays, then processes incoming messages one after the other; and (iii) the dynamic system and measurement models are linear with additive white Gaussian noise (AWGN).
Note that we choose to use the CF for its conceptual simplicity, as it allows us to directly compute the fused marginal pdf, $p_f(\chi_C^{ij})$, in (\ref{eq:BDF_fusion}) and (\ref{eq:HeteroFusion_3states_i}), and as a way to gain fundamental understanding into the heterogeneous fusion problem. However, the algorithm presented is general in the sense that different methods to explicitly track or approximate $p^{ij}_c$ can be used when assumptions (i)-(iii) don't necessarily hold, e.g., by pedigree tracking \cite{martin_distributed_2005} or GMD approximation \cite{bailey_conservative_2012}, respectively.

\textbf{The Gaussian information form:}
The information form of the Gaussian distribution is particularly convenient for deriving and describing key steps in data fusion processing, for example in the information filter, the CF \cite{grime_data_1994}, the CI algorithm \cite{julier_non-divergent_1997}, and more.
This allows to directly `read' conditional independence from the information matrix, develop closed-form heterogeneous fusion rules, namely summing and subtracting the sufficient statistics (information vector and information matrix), and to suggest approximations for conservative marginalization, where the term `conservative' is defined in clear terms.

\subsection{Preliminaries}
Assume the full joint distribution over the random state vector $\chi$, is a multivariate Gaussian with mean $\mu$ and covariance matrix $\Sigma$. 
\begin{comment}
\begin{equation}
    \mu=\begin{pmatrix}
    \mu_{\chi_C}\\
    \mu_{\chi_L}\\
    \end{pmatrix}, \ \ \ \ \ \ \ \Sigma=\covMat  \\
\end{equation}
\end{comment}
The pdf in information form for the normally distributed state $\chi$, with information vector $\zeta$ and information matrix $\Lambda$ is \cite{schon_manipulating_2011}:
\begin{equation}
    p(\chi;\zeta,\Lambda) = \frac{\exp(-\frac{1}{2}\zeta^T\Lambda^{-1}\zeta)}{\det(2\pi\Lambda^{-1})^\frac{1}{2}} \exp\big({-\frac{1}{2}\chi^T\Lambda \chi+\zeta^T\chi\big)},
    \label{eq:infoDist}
\end{equation}
with, 
\begin{equation}
    \zeta=\Sigma^{-1}\mu=\begin{pmatrix}
    \zeta_{\chi_C}\\
    \zeta_{\chi_L}\\
    \end{pmatrix}, \ \  \Lambda=\Sigma^{-1}=\infoMat.  \\
    \label{eq:infoDef}
\end{equation}
Here $\chi_C$ and $\chi_L$ are the common and local subsets of random states, respectively.
This pdf can also be expressed using factorization (\ref{eq:factorDist}), where the marginal and conditional distributions of a Gaussian are also Gaussian, 
\begin{equation}
    \begin{split}
         p(\chi_C)&=\mathcal{N}^{-1}(\chi_C;\bar{\zeta}_{\chi_C},\bar{\Lambda}_{\chi_C\chi_C}) \\
         p(\chi_L|\chi_C)&=\mathcal{N}^{-1}(\chi_L;\zeta_{\chi_L|\chi_C},\Lambda_{\chi_L|\chi_C}) 
         \label{eq:pdfDef}
    \end{split}
\end{equation}
with $\mathcal{N}^{-1}$ representing the information form of the Gaussian distribution $\mathcal{N}$, and $(\bar{\zeta}_{\chi_C},\bar{\Lambda}_{\chi_C\chi_C})$ and $(\zeta_{\chi_L|\chi_C},\Lambda_{\chi_L|\chi_C})$ are the sufficient statistics for the marginal and conditional pdfs in information form, respectively, defined as \cite{thrun_graphslam_2005}:
 \begin{equation}
     \begin{split}
         &\bar{\zeta}_{\chi_C} = \zeta_{\chi_C}-\Lambda_{\chi_C\chi_L}\Lambda^{-1}_{\chi_L\chi_L}\zeta_{\chi_L},\\ &\bar{\Lambda}_{\chi_C\chi_C}=\Lambda_{\chi_C\chi_C}-\Lambda_{\chi_C\chi_L}\Lambda^{-1}_{\chi_L\chi_L}\Lambda_{\chi_L\chi_C} \\
         &\zeta_{\chi_L|\chi_C} = \zeta_{\chi_L}-\Lambda_{\chi_L\chi_C}\chi_C, \ \  \Lambda_{\chi_L|\chi_C}=\Lambda_{\chi_L\chi_L}
         \label{eq:marg_cond}
     \end{split}
 \end{equation}

\subsection{Fusion}
To develop the closed form heterogeneous CF algorithms we start with the original homogeneous fusion rule (\ref{eq:bayesCF}). By substituting linear Gaussian distributions, taking logs and differentiating once for the fused information vector ($\zeta_f$) and twice for the fused information matrix ($\Lambda_f$), \cite{chong_distributed_1983} the following fusion equations can be obtained,
\begin{equation}
    \begin{split}
        \zeta_f =\zeta^i +\zeta^j - \zeta_c^{ij}\ , \ \ \ \ \
        \Lambda_f=\Lambda^i+\Lambda^j-\Lambda_c^{ij}.
    \end{split}
    \label{eq:CFequation}
\end{equation}
These equations are the basis of the original linear-Gaussian CF \cite{grime_data_1994}, which explicitly tracks the `common information' vector and matrix $(\zeta_c^{ij}, \Lambda_c^{ij})$, describing the pdf over $\chi$ conditioned on the common data between communicating pairs of agents $i$ and $j$ ($p_c^{ij}(\chi|Z^{i,-}\cap Z^{j,-})$) in an undirected acyclic communication graph.

Define $\Bar{\zeta}_{\chi_C^{ij},f}$ and $\Bar{\Lambda}_{\chi_C^{ij}\chi_C^{ij},f}$ to be the \emph{fused} marginal information vector and matrix, respectively, over the common random state $\chi_C^{ij}$ between agents $i$ and $j$, corresponding to $p_f(\chi_C^{ij}|Z^+)$, represented in information form.
Without loss of generality, the fused marginal information vector and matrix can be achieved using different fusion methods, exact and approximate (CI \cite{julier_non-divergent_1997} for example).
This paper restricts attention to the CF for exact fusion. 
Then by using (\ref{eq:CFequation}), the fused marginal information vector and matrix for Gaussian distributions are given by

\begin{equation}
    \begin{split}
        &\bar{\zeta}_{\chi_C^{ij},f} = \bar{\zeta}^i_{\chi_C^{ij}}+\bar{\zeta}^j_{\chi_C^{ij}}-\bar{\zeta}_{\chi_C^{ij},c}^{ij},\\ 
        &\bar{\Lambda}_{\chi_C^{ij}\chi_C^{ij},f} = \bar{\Lambda}^i_{\chi_C^{ij}\chi_C^{ij}}+\bar{\Lambda}^j_{\chi_C^{ij}\chi_C^{ij}}-\bar{\Lambda}^{ij}_{\chi_C^{ij}\chi_C^{ij},c}.
        \label{eq:GaussMarginal}
    \end{split}
\end{equation}

The HS-fusion rule (\ref{eq:HeteroFusion_3states_i}) for Gaussian pdfs is now dubbed \emph{HS-CF} and is represented by the simple closed form expression,

\begin{equation}
    \begin{split}
        \zeta^{i}_f=&\left( \begin{array}{c}
             \bar{\zeta}_{\chi_C^{ij},f}   \\ \hdashline[2pt/2pt]
             0
        \end{array} \right)
        +\left( \begin{array}{c}
              \Lambda^i_{\chi_C^{ij}\chi_L^i}(\Lambda^{i}_{\chi_L^i\chi_L^i})^{-1}\zeta^i_{\chi_L^i} \\ \hdashline[2pt/2pt]
              \zeta^i_{\chi_L^i}
        \end{array}\right)  \\
        \Lambda^{i}_f=&\left(\begin{array}{c;{2pt/2pt}c}
             \bar{\Lambda}_{\chi_C^{ij}\chi_C^{ij},f}  & 0  \\ \hdashline[2pt/2pt]
             0 & 0 \end{array} \right)
             + \\ &\left(\begin{array}{c;{2pt/2pt}c}
             \Lambda^i_{\chi_C^{ij}\chi_L^i}(\Lambda^i_{\chi_L^i\chi_L^i})^{-1}\Lambda^i_{\chi_L^i\chi_C^{ij}}  & \Lambda^i_{\chi_C^{ij}\chi_L^i}  \\
             \hdashline[2pt/2pt]
             \Lambda^i_{\chi_L^i\chi_C^{ij}} & \Lambda^i_{\chi_L^i\chi_L^i} \end{array} \right),
    \label{eq:fusionEq}
    \end{split}
\end{equation}
where $\bar{\zeta}_{\chi_C^{ij},f}$ and $\bar{\Lambda}_{\chi_C^{ij}\chi_C^{ij},f}$ are given in (\ref{eq:GaussMarginal}) and $\chi_L^i\in\chi_L$ is the subset of $i$'s local random states. An equivalent expression for the fused information vector and matrix at agent $j$ is achieved by switching $i$ with $j$. 

It is important to note, as seen from (\ref{eq:GaussMarginal}), that the fused marginal pdf  $p(\chi_C^{ij}|Z^+)=\mathcal{N}^{-1}(\bar{\zeta}_{\chi_C^{ij},f},\bar{\Lambda}_{\chi_C^{ij}\chi_C^{ij},f})$ is the same for agents $i$ and $j$. However, the conditional part is kept local (the right part of (\ref{eq:fusionEq})), which means that after fusion, the \emph{local joint} distributions in $i$ (w.r.t. $\chi^i=\chi_C^{ij}\cup\chi_L^i$) and $j$ (w.r.t. $\chi^j=\chi_C^{ij}\cup\chi_L^j$) are different. While agents only update the information vector and matrix of the marginal pdf, over the common random state $\chi_C^{ij}$, the local joint distribution in moment representation (e.g., $\mu^{i}_f, \Sigma^{i}_f$) will be updated, thus also updating the local states $\chi_L^i$ ($\chi_L^j$).

\subsection{The Information Augmented State Smoother}
\label{subSec:iAS}
%\subsection{The Information Augmented State}
%\label{subSec:iAS}

To maintain conditional independence in dynamic systems, each agent has to hold a distribution over the augmented state $p(\chi_{k:0}|Z_k^-)$ given in (\ref{eq:pdf_aug_state}).  
For Gaussian distributions there are several versions to recursively augment the state. The \emph{augmented} state (AS) \cite{chong_comparison_2014}, and the \emph{accumulated} state density (ASD) \cite{koch_accumulated_2011} are similar, but not equivalent, where they mostly differ in their retrodiction formulation. Reference \cite{chong_comparison_2014} uses covariance formulation for the prediction step, and information representation for the update step. However, since the algorithms developed in this paper work in Gaussian information space, it is advantageous to work with a full information filter formulation. The authors in \cite{eustice_exactly_2005} derive a nonlinear version of the augmented state filter, which they name `delayed-state', approximated by the first two moments, similar to the extended information filter (EIF). 
To be able to use the augmented state formulation, we provide here a linear-Gaussian version of \cite{eustice_exactly_2005}, which we dub the \emph{Information Augmented State} (\emph{iAS}) smoother. 

Consider a dynamic system, described by the linear discrete time equations
\begin{equation}
    \begin{split}
        &x_k=F_kx_{k-1}+Gu_k+\omega_k,  \ \ \  \ \omega_k\sim \mathcal{N}(0,Q_k)\\
        &y_k=H_kx_k+v_k,  \ \ \ \ \ \ \ \ \ \ \ \ \ \ \ \ v_k\sim \mathcal{N}(0,R_k),
        \label{eq:dynamicSys}
    \end{split}
\end{equation}
where $F_k$ is the state transition matrix, $G$ is the control effect matrix and $H_k$ is the sensing matrix. $\omega_k$ and $v_k$ are the zero mean white Gaussian process and measurement noise, respectively.

The predicted information vector and matrix for the time window $k:n$, given all information up to and including time step $k-1$ are given by
\begin{equation}
    \begin{split}
     &\zeta_{k:n|k-1}=\begin{pmatrix}Q_k^{-1}Gu_k \\ 
      \zeta_{k-1:n|k-1}-\mathbf{F}^TQ_k^{-1}Gu_k\end{pmatrix}\\
       &\Lambda_{k:n|k-1}=
       \begin{pmatrix} Q_k^{-1} & -Q_k^{-1}\mathbf{F}  \\
       -\mathbf{F}^TQ_k^{-1}  & \Lambda_{k-1:n|k-1}+\mathbf{F}^TQ_k^{-1}\mathbf{F}
        \end{pmatrix},\\
    \end{split}
        \label{eq:iAS_pred_vec}
\end{equation}
where $\mathbf{F}=\big[F_{k-1} \ \ 0_{m \times m(k-n-2)} \big ]$ and $m$ is the size of the (not augmented) state vector.
\begin{comment}
Notice the simplicity of the above expression and its interesting interpretation: the predicted conditional information matrix at the current time step, given previous time steps (upper left block), depends inversely on process noise alone and is not affected by the state dynamics. 
\end{comment}

For completeness, the measurement update in Gaussian information space is \cite{chong_comparison_2014}
\begin{equation}
    \begin{split}
       &\zeta_{k:n|k}= \zeta_{k:n|k-1}+J_ki_k \\
       &\Lambda_{k:n|k}= \Lambda_{k:n|k-1}+J_kI_kJ_k^T,
        \label{eq:iAS_upd_vec}
    \end{split}
\end{equation}
where $J_k=\big[I_m \ \ 0_{m \times m(k-n-1)} \big ]^T$, $i_k=H_k^TR_k^{-1}y_k$ and $I_k=H_k^TR_k^{-1}H_k$.

In linear-Gaussian problems, the above \emph{iAS} can be used locally at each agent $i$, enabling conditional independence such that for $n=1$,
\begin{equation}
    \begin{split}
    p(\chi_{C,k:1},\chi_L^i,\chi_L^j&|Z_k^{i,+}) = p(\chi_{C,k:1}|Z_k^{i,+})\cdot \\ p(&\chi_L^i|\chi_{C,k:1},Z_k^{i,-})\cdot p(\chi_L^j|\chi_{C,k:1},Z_k^{j,-}).
    \end{split}
\end{equation}
It can be seen that the full information matrix and vector given in (\ref{eq:iAS_pred_vec}) grows rapidly with the time step $k$, inducing high processing and communication costs. 
While the block tri-diagonal structure of the updated information matrix $\Lambda_{k:1|k-1}$, resulting from the Markov property of the system dynamics can be utilized to reduce computation burden, it does not resolve the communication load problem which scales with the size of the network - as messages need to propagate through from one end of the network graph to the other. 
Instead, a filtering approach is taken to marginalize past states to process only a sliding window $k:n$ ($n>1$) while maintaining conditional independence.
This requires conservative filtering, discussed in the next section.

\subsection{Conservative Filtering}
%\subsection{Conservative Sparse Marginalization}
\label{subSec:GaussianSparseMarg}
%\label{sec:consFilter} 

%To keep the paper succinct we focus our attention on the Bi-directional factorized CF (BDF-CF) and the Heterogeneous state CF (HS-CF), but the methods derived bellow can be easily extended to other CF$^2$ algorithms. 

%In all CF$^2$ algorithms except the HS-CF we assume all agents hold an estimated distribution over the same full set of states. 
%while in HS-CF the estimated distribution is over different but overlapping subsets. This requires different treatment to local filtering, more specifically - in prediction step where previous dynamic states (target position in our problem) are marginalized out. Fig. \ref{fig:GraphModel2} is used to demonstrate the problem by looking at the local distribution at any agent $i$.
%The challenges of dynamic problems are addressed differently for the BDF-CF and the HS-CF. 
 %(\ref{subSec:GaussianSparseMarg}). 
\begin{comment}

\begin{figure}[tb]
      \centering
      %\framebox{\parbox{3in}{}
      \includegraphics[scale=0.4]{Figures/GraphModel2_v2.eps}
      \caption{\textcolor{red}{Need to insert a new graph}}
      \label{fig:GraphModel2}
      \vspace{-0.2in}
\end{figure}
\end{comment}

In the graphSLAM literature the idea of disconnecting dependencies between landmarks due to marginalization of past robot states is known as conservative sparsification \cite{vial_conservative_2011,carlevaris-bianco_conservative_2014}. A dense information matrix is approximated by a sparse one, with the goal of reducing the \emph{computational complexity} when reasoning over the map. The approach of Vial \etal \cite{vial_conservative_2011} is adopted here with the goal of enforcing \emph{conditional independence} after marginalization of past states.

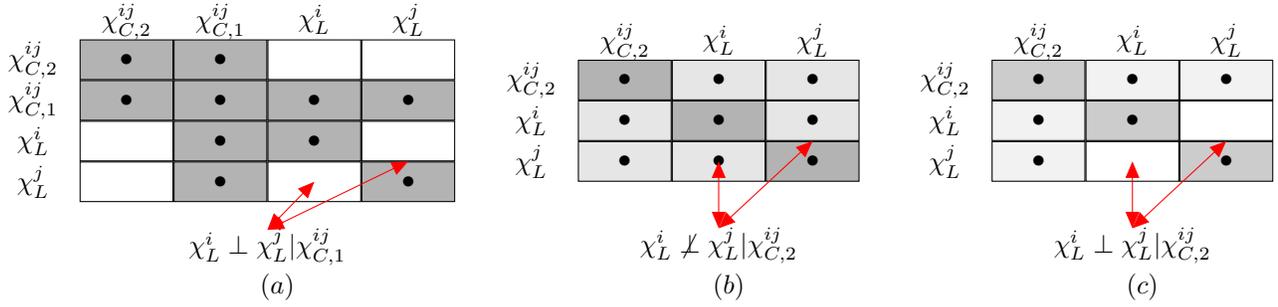
\begin{figure*}[tb!]
\begin{tikzpicture}

    \node (a) at (0.75,-2.25) {$(a)$}; 
    \matrix [
    matrix of nodes, nodes in empty cells, 
    nodes={text width=1cm, align=center,
    minimum height=1.5em, anchor=center, draw},
    column 1/.style={ % Override stripes and modify the feature column
    {nodes={fill=none, draw=none}},
    nodes={inner ysep=0},},
% modify headers first via common styles and then specific colors
    row 1/.style={nodes={text depth=0.2ex, text width=0.75cm, fill=none, draw=none}}, 
    row 6/.style={nodes={text depth=-1cm, fill=none, draw=none}}, 
    ] (m1) at (0,0)
    {
    & $\chi_{C,2}^{ij}$ & $\chi_{C,1}^{ij}$ & $\chi_{L}^{i}$ & $\chi_{L}^{j}$\\
     $\chi_{C,2}^{ij}$& \node[fill=black!30] {$\bullet$}; & \node[fill=black!30] {$\bullet$}; &  &  \\
    $\chi_{C,1}^{ij}$& \node[fill=black!30] {$\bullet$}; & \node[fill=black!30] {$\bullet$}; & \node[fill=black!30] {$\bullet$}; & \node[fill=black!30] {$\bullet$}; \\
    $\chi_{L}^{i}$  & & \node[fill=black!30] {$\bullet$}; & \node[fill=black!30] {$\bullet$}; & \node (a) {}; \\
    $\chi_{L}^{j}$& & \node[fill=black!30] {$\bullet$}; & \node (b) {}; & \node[fill=black!30] {$\bullet$}; \\
    & & & &\\
    };
    
    \node (eq)[fit=(m1-6-3)(m1-6-4), yshift=-0.5cm, draw=none]{$\chi_{L}^i\perp \chi_L^j|\chi^{ij}_{C,1}$};
    \draw [<->,red,thin] (eq.north) -- (a.south) node [] {};
    \draw [<->,red,thin] (eq.north) -- (b.center) node [] {};

    \node (b) at (6.75,-2.25) {$(b)$};
    \matrix [
    matrix of nodes, nodes in empty cells, 
    nodes={text width=1cm, align=center,
    minimum height=1.5em, anchor=center, draw},
    column 1/.style={ % Override stripes and modify the feature column
    {nodes={fill=none, draw=none}},
    nodes={inner ysep=0},},
% modify headers first via common styles and then specific colors
    row 1/.style={nodes={text depth=0.2ex, text width=0.75cm, fill=none, draw=none}}, 
    row 5/.style={nodes={text depth=-1cm, fill=none, draw=none}}, 
    %head color/.list={2/orange,3/teal,4/cyan,5/magenta}
    ] (m2) at (6,0)
    {
    & $\chi_{C,2}^{ij}$ & $\chi_{L}^{i}$ & $\chi_{L}^{j}$\\
     $\chi_{C,2}^{ij}$& \node[fill=black!30] {$\bullet$}; & \node[fill=black!10] {$\bullet$}; &  \node[fill=black!10] {$\bullet$};   \\
    $\chi_{L}^{i}$  &  \node[fill=black!10] {$\bullet$};& \node[fill=black!30] {$\bullet$}; & \node (a1) [fill=black!10] {$\bullet$};   \\
    $\chi_{L}^{j}$& \node[fill=black!10] {$\bullet$}; & \node (b1) [fill=black!10] {$\bullet$}; &  \node[fill=black!30] {$\bullet$}; \\
    & & & \\
    };
    \node (eq1)[fit=(m2-5-2)(m2-5-4), yshift=-0.75cm, draw=none]{$\chi_{L}^i\not\perp \chi_L^j|\chi^{ij}_{C,2}$};
    \draw [<->,red,thin] (eq1.north) -- (a1.south) node [] {};
    \draw [<->,red,thin] (eq1.north) -- (b1.center) node [] {};
    
    \node (c) at (12.25,-2.25) {$(c)$};
    \matrix [
    matrix of nodes, nodes in empty cells, 
    nodes={text width=1cm, align=center,
    minimum height=1.5em, anchor=center, draw},
    column 1/.style={ % Override stripes and modify the feature column
    {nodes={fill=none, draw=none}},
    nodes={inner ysep=0},},
% modify headers first via common styles and then specific colors
    row 1/.style={nodes={text depth=0.2ex, text width=0.75cm, fill=none, draw=none}}, 
    row 5/.style={nodes={text depth=-1cm, fill=none, draw=none}},
    %head color/.list={2/orange,3/teal,4/cyan,5/magenta}
    ] (m3) at (11.5,0)
    {
    & $\chi_{C,2}^{ij}$ & $\chi_{L}^{i}$ & $\chi_{L}^{j}$\\
     $\chi_{C,2}^{ij}$& \node[fill=black!20] {$\bullet$}; & \node[fill=black!5] {$\bullet$}; &  \node[fill=black!5] {$\bullet$};   \\
    $\chi_{L}^{i}$  &  \node[fill=black!5] {$\bullet$};& \node[fill=black!20] {$\bullet$}; & \node (a2) {};  \\
    $\chi_{L}^{j}$& \node[fill=black!5] {$\bullet$}; & \node (b2) {}; &  \node[fill=black!20] {$\bullet$}; \\
    & & & \\
    };
    \node (eq2)[fit=(m3-5-2)(m3-5-4), yshift=-0.75cm, draw=none]{$\chi_{L}^i\perp \chi_L^j|\chi^{ij}_{C,2}$};
    \draw [<->,red,thin] (eq2.north) -- (a2.south) node [] {};
    \draw [<->,red,thin] (eq2.north) -- (b2.center) node [] {};

\end{tikzpicture}
    
\caption{Information matrix visualisation showing conditional independence. (a) Before marginalization of time step $1$, local rv subsets are conditionally independent given common rv subsets, indicated by empty (zero) cells in the matrix. (b) Marginalizing $\chi_{C,1}^{ij}$ results in filling in the matrix or direct dependencies between local rv subsets. (c) Conservative filtering regains conditional independence by setting matrix cells to zero and deflating the matrix. }
    \label{fig:infoMatrix}
    \vspace{-0.2in}
\end{figure*}

The sparse structure of the marginalized approximate information matrix is enforced by removing the links between $\chi_L^i$ and $\chi_L^j$. In other words, given a true dense Gaussian distribution $\mathcal{N}(\zeta_{tr},\Lambda_{tr})$, a sparse approximate distribution $\mathcal{N}(\zeta_{sp},\Lambda_{sp})$ is sought such that the mean is unchanged and the approximation is conservative in the PSD sense,
\begin{equation}
    \begin{split}
        \Lambda_{tr}^{-1}\zeta_{tr}=\Lambda_{sp}^{-1}\zeta_{sp},\ \ \ \ \
        \Lambda_{tr}-\Lambda_{sp}\succeq 0,
    \end{split}
    \label{eq:consFilter}
\end{equation}
where again the information form of the Gaussian distribution is used. A visualization of the information matrix is given in Fig. \ref{fig:infoMatrix}, where bulleted-filled cells mean non-zero terms and empty cells mean zero terms in the information matrix, with the latter indicating conditional independence. Fig. \ref{fig:infoMatrix}(b) shows that filtering causes direct dependencies between local rv subsets, indicated by the filling of corresponding cells in the information matrix. Then, we wish to regain conditional independence by setting matrix terms to zero, or by `emptying' cells as shown in Fig. \ref{fig:infoMatrix}(c). To ensure conservativeness the information matrix has to be deflated, shown by faded gray color of the cells. Note that since an agent $i$ in HS-fusion only estimate their own local states ($\chi_L^i$) and not their neighbor's ($\chi_L^j$), the suggested conservative marginalization method only applies to BDF-fusion.

Reference \cite{vial_conservative_2011}  minimizes the Kullback-Leibler Divergence (KLD) to find a lower bound on the dense true information matrix $\Lambda_{tr}$. 
Along similar lines, \cite{forsling_consistent_2019} suggests a method named `Uniform Pre-Transmission Eigenvalue-Based Scaling' to conservatively approximate a covariance matrix $\Sigma$ by inflating a diagonal matrix $D$ built out of the diagonal entries of the full matrix $\Sigma$. 
To achieve a conservative approximation $D_c$, $D$ is inflated by multiplying by the largest eigenvalue of $Q=D^{-\frac{1}{2}}\Sigma D^{-\frac{1}{2}}$. 
This results in $D_c=\lambda_{max}D$ such that $D_c-\Sigma \succeq 0$.

This method is generalized here to find a lower bound sparse information matrix $\Lambda_{sp}$ and regain conditional independence between $\chi_L^i$ and $\chi_L^j$. 
This new generalized method differs from the one suggested in \cite{forsling_consistent_2019} in two ways.
Firstly, the approximation, $\Lambda_{sp}$, is allowed to be \emph{any} information matrix achieved by setting any off-diagonal elements of the true dense information matrix $\Lambda_{tr}$ to zero, i.e., the resulting matrix is in general not diagonal or even block-diagonal (e.g., as in Fig. \ref{fig:infoMatrix}(c)). 
Note that since the information matrix (i.e. not the covariance) is changed, setting off-diagonal elements to zero directly controls the conditional independence structure of the underlying distribution. 
Specifically for the purpose of this paper, terms relating local random states (e.g., $\chi_L^i$ and $\chi_L^j$) in $\Lambda_{tr}$ are set to zero to regain conditional independence given common states (e.g., $\chi_{C,k:n}^{ij}$).

The second change from the original method is that the information matrix is approximated, as opposed to the covariance matrix. This means that a \emph{lower} bound is sought and not an \emph{upper} bound, i.e. `information' must be deflated, instead of uncertainty being inflated. 
This is achieved by choosing the \emph{minimal} eigenvalue of $\Tilde{Q}=\Lambda_{sp}^{-\frac{1}{2}}\Lambda_{tr}\Lambda_{sp}^{-\frac{1}{2}}$, resulting in
\begin{equation}
    \Lambda_{tr}-\lambda_{min}\Lambda_{sp}\succeq 0,
\end{equation}
where $\lambda_{min}\Lambda_{sp}$ is the sought of conservative marginal sparse approximation of the dense information matrix $\Lambda_{tr}$.
The new information vector is computed to such that (\ref{eq:consFilter}) holds
\begin{equation}
    \zeta_{sp} =(\lambda_{min}\Lambda_{sp})\Lambda_{tr}^{-1}\zeta_{tr}.
\end{equation}

\subsection{Closed-Form Algorithm}
This subsection provides a full summary description of the different steps each agent $i$ takes to locally process, communicate and fuse data with its neighbor $j$. 
While the steps described in the pseudo code in Algorithm \ref{algo:CF2} are general in the sense that it can be applied with any pdf, it assumes linear-Gaussian distributions and uses the theory developed above to provide equation references (green arrows) for the closed-form expressions of the different operations.
The algorithms are named \emph{BDF-CF} and \emph{HS-CF}, as the agents use the CF to track dependencies in the data and are used as an example to detail the algorithm from the perspective of one agent $i$ communicating with a neighbor $j$. 

\setlength{\textfloatsep}{0.25in}
\begin{algorithm}
    \caption{BDF-CF / HS-CF algorithm}
    \label{algo:CF2}
    \begin{algorithmic}[1]
    \State {Define: $\chi_{i}$, $\chi_{j}$, $\chi_{c}^{ij}$, $\chi_{\neg i}^{ij}$, $\chi_{\neg j}^{ji}$, prior pdfs} \Comment{{\color{ForestGreen} Fig. \ref{fig:setDiagram}}}
    \For{All time steps}
    \State Propagate state local states \Comment{{\color{ForestGreen} Eq. \ref{eq:iAS_pred_vec}}}
    \State Propagate common states in the CF
    \If{BDF-CF}
    \State Conservative filtering \Comment{{\color{ForestGreen} Sec. \ref{subSec:GaussianSparseMarg}}}
    \ElsIf{HS-CF}
    \State Marginalize out past state
    \EndIf 
    \State Measurement update \Comment{{\color{ForestGreen} Eq. \ref{eq:iAS_upd_vec}}}
    \State Send message \Comment{{\color{ForestGreen} Eq. \ref{eq:message}}} 
    \State Fuse received message \Comment{{\color{ForestGreen} Eq. \ref{eq:setFusionPDF}}}  
    \State Update common information
    \EndFor
    \State \Return
    \end{algorithmic}
\end{algorithm}

\subsection{Calculation of Communication and Computation Savings}
\label{sec:NumExample} 

To highlight the potential gain of the proposed heterogeneous fusion rules
with respect to communication and computation complexity and how they change with scale, three numerical examples (small, medium and large) of a multi-agent multi-target tracking problem are presented.  
Consider the problem introduced earlier of tracking $n_t$ ground targets by $n_a$ agents (trackers), where each agent computes a local KF estimate, i.e., the system dynamics are assumed to be linear with additive Gaussian white noise (\ref{eq:dynamicSys}). 
Each agent $i\in N_a$ has 6 unknown local position states described by the random vector $\chi_L^i=s^i$ and takes measurements to $n_t^i$ targets, each having 4 position/velocity states described by the random vector $x_t$ (e.g., east and north coordinates and velocities). 
The full state vector then has $6n_a+4n_t$ random states. 
Assume a tree topology in ascending order, where each agent tracks 1/2/3 targets, corresponding to the small/medium/large examples, respectively, but has only one target in common with its neighbor.
Now, using the same logic as before, assume that each agent is only concerned with the targets it takes measurements of, and its own position states. 
Each agent has only 10/14/18 local `relevant' random states for tracking 1/2/3 targets, respectively.

Table \ref{tab:table2} presents a comparison between the different heterogeneous fusion algorithms, when a channel filter is used (Algorithm \ref{algo:CF2}), for the three different scale problems, in a filtering scenario. 
The baseline for comparison is the original (homogeneous) CF, with each agent estimating the full state vector. 
For the communication data requirement, double precision (8 bytes per element) is assumed. 
Since the matrices are symmetric covariances, agents only need to communicate $n(n+1)/2$ upper diagonal elements, where $n$ is the number of random states.
Each agent's computation complexity is determined by the cost of inverting an $n\times n$ matrix. 
It can be seen from the table that
even for the small scale problem, the communication data reduction is significant; the BDF-CF requires about 42.7\% of the original CF, while the HS-CF requires only 9.2\% as agents only communicate common targets states information vectors and matrices. 
These gains then increase with scale, for the medium and large problems the BDF-CF communication is about 33\% of the original CF and for the HS-CF is less than 1\%.

Another important aspect is computation complexity. 
As seen from the table, while the BDF-CF algorithm requires each agent to process the full random state vector, the HS-CF offers significant computational reduction. 
Since each agent only processes the locally relevant random states, the HS-CF scales with subset of states and not with number of agents and targets. %We can see that 
In the medium and large scale problems, as the size of the full system random vector states increase, the HS-CF computation complexity is less than 1\% of the other algorithms, which can be critical in terms of computing power for resource-constrained platforms.

% Dynamic example
\begin{comment}

\begin{table}[tb]
\caption{Data communication requirements and computational complexity for different fusion methods}
%\label{table_example}
\begin{center}
\begin{tabular}{c c c c}
  {\bf Method} & {\bf Data [bytes]  }& {\bf Data [\%CF]  }  & {\bf Complexity}\\ 
\hline
  {CF}   & 801216     & ---  & $O(104^3)$  \\
\hline
  {BDF-CF}   & 273216     & 34.1  & $O(104^3)$  \\
\hline
  {Approximate BDF-CF}   & 3456     & 0.43  & $O(104^3)$  \\
\hline
  {HS-CF}   & 3456     & 0.43  & $O(18^3)$  \\
\end{tabular}
\end{center}
\label{tab:table2}
%\vspace{-0.25in}
\end{table}

% new table
\begin{table}[tb]
\caption{Data communication requirements and computational complexity for different fusion methods}
%\label{table_example}
\begin{center}
\begin{tabular}{c c c c}
   & {\bf Small  }& {\bf Medium }  & {\bf Large}\\ 
\hline
  {(Agents, Targets)}   & (2, 1)     & (10, 11)  & (25, 51)  \\
\hline
  Targets per agent   & 1     & 2  & 3  \\
\hline
  {CF data req. [KB]}   & 2.4     & 801  & 24300  \\
\hline
  {CF complexity}   & $O(16^3)$  & $O(104^3)$  & $O(354^3)$  \\
\hline
  {BDF-CF data req. [\%CF] }  & 42.7  & 33.7  & 33.2\\
\hline
  {BDF-CF complexity [\%CF] }  & 100  & 100  & 100\\
\hline
  {Approx. BDF-CF data req. [\%CF] }    & 9.2  & 0.25  & 0.02\\
\hline
  {Approx. BDF-CF complexity [\%CF] }    & 100  & 100  & 100\\
\hline
  {HS-CF data req. [\%CF] }   & 9.2  & 0.25  & 0.02\\
\hline
  {HS-CF complexity [\%CF] }   & 24.4  & 0.24  & 0.01\\
\end{tabular}
\end{center}
\label{tab:table2}
%\vspace{-0.25in}
\end{table}
\end{comment}

\begin{table}[tb]
\caption{Data communication requirements and computational complexity for different fusion algorithms, for different problem scales.}
\renewcommand{\arraystretch}{1.5}
\centering
\begin{tabular}{llccc}
&& Small     & Medium     & Large       \\ 
\hline
  & ($n_a$, $n_t$) & (2, 1)    & (10, 11)   & (25, 51)    \\ 
\cline{2-5}
& $n_t^i$ & 1         & 2          & 3           \\ 
\hline \hline
\multirow{2}{*}{CF}                                                       & Data req. $[$KB$]$    & 2.4       & 801        & 24300       \\ 
\cline{2-5}
& Complexity        & $O(16^3)$ & $O(104^3)$ & $O(354^3)$  \\ 
\hline \hline
\multirow{2}{*}{BDF-CF}                                                   & Data req. $[$\%CF$]$  & 42.7      & 33.7       & 33.2        \\ 
\cline{2-5}
        &Complexity  &  $O(16^3)$ & $O(104^3)$ & $O(354^3)$         \\ 
%\hline \hline
%\multirow{2}{*}{\begin{tabular}[c]{@{}l@{}}Approx. \\BDF-CF\end{tabular}} &Data req. $[$\%CF$]$  & 9.2       & 0.25       & 0.02        \\ 
%\cline{3-5}
%    & Complexity  & $O(16^3)$ & $O(104^3)$ & $O(354^3)$         \\ 
\hline \hline
\multirow{2}{*}{HS-CF}                                                    & Data req. $[$\%CF$]$ & 9.2       & 0.25       & 0.02        \\
       & Complexity & $O(10^3)$ & $O(14^3)$ & $O(18^3)$    
\end{tabular}
\label{tab:table2}
\end{table}

% Table without BDF-CF
% \begin{table}[tb]
% \caption{Data communication requirements and computational complexity for different fusion algorithms, for different problem scales.}
% \renewcommand{\arraystretch}{1.5}
% \centering
% \begin{tabular}{llccc}
% && Small     & Medium     & Large       \\ 
% \hline
%   & ($n_a$, $n_t$) & (2, 1)    & (10, 11)   & (25, 51)    \\ 
% \cline{2-5}
% & $n_t^i$ & 1         & 2          & 3           \\ 
% \hline \hline
% \multirow{2}{*}{CF}                                                       & Data req. $[$KB$]$    & 2.4       & 801        & 24300       \\ 
% \cline{2-5}
% & Complexity        & $O(16^3)$ & $O(104^3)$ & $O(354^3)$  \\ 
% % \hline \hline
% % \multirow{2}{*}{BDF-CF}                                                   & Data req. $[$\%CF$]$  & 42.7      & 33.7       & 33.2        \\ 
% % \cline{2-5}
% %         &Complexity  &  $O(16^3)$ & $O(104^3)$ & $O(354^3)$         \\ 
% %\hline \hline
% %\multirow{2}{*}{\begin{tabular}[c]{@{}l@{}}Approx. \\BDF-CF\end{tabular}} &Data req. $[$\%CF$]$  & 9.2       & 0.25       & 0.02        \\ 
% %\cline{3-5}
% %    & Complexity  & $O(16^3)$ & $O(104^3)$ & $O(354^3)$         \\ 
% \hline \hline
% \multirow{2}{*}{HS-CF}                                                    & Data req. $[$\%CF$]$ & 9.2       & 0.25       & 0.02        \\
%        & Complexity & $O(10^3)$ & $O(14^3)$ & $O(18^3)$    
% \end{tabular}
% \label{tab:table2}
% \end{table}

\vspace{-0.05in}
\section{Simulation Studies}
\label{sec:Sim}
Multi-agent multi-target tracking simulation scenarios where performed to compare and validate the proposed algorithms. 
Since the dynamics and measurement models are assumed to be linear with Gaussian noise, Algorithm \ref{algo:CF2} is used together with the \emph{iAS} as the inference engine, i.e., agents estimate the sufficient statistics (information vector and matrix) of the random state vector. 
First, the algorithms are tested on a static target case, where conditional independence of the local states can be easily guaranteed. 
This is followed by a dynamic target test case with only two agents and one target, to validate and compare the smoothing and the conservative filtering approaches. 
Lastly, the conservative filtering approach is used for a more challenging 4-agents 5-target scenario.
Results for all the different scenarios are based on Monte Carlo simulations and compare the new algorithms to an optimal centralized estimator.

\subsection{Example 1 - Static Case}
A chain network, consisting of five agents connected bidirectionally in ascending order $(1\leftrightarrow 2\leftrightarrow 3\leftrightarrow 4\leftrightarrow 5)$, as depicted in (Fig. \ref{fig:targetTrackingExample}), attempts to estimate the position of six stationary targets in a $2D$ space.
Assume each tracking agent $i\ =1,...,5$ has perfect self position knowledge, but with a constant agent-target relative position measurement bias vector in the east and north directions $s^i=[b^i_{e},b^i_{n}]^T$. 
In every time step $k$, each agent takes two kinds of measurements: one for the target and one to collect data on the local sensor bias random vector, which can be transformed into the  linear pseudo-measurements,
\begin{align}
    \begin{split}
        y^{i,t}_{k} &= x^t+s^i+v^{i,1}_k, \ \ v^{i,1}_k \sim \mathcal{N}(0,R^{i,1}),  \\
        m^i_{k} &= s^i+v^{i,2}_k, \ \ v^{i,2}_k \sim \mathcal{N}(0,R^{i,2}),
    \end{split}
    \label{eq:measModel}
\end{align}
where $y^{i,t}_{k}$ is agent $i$'s relative measurement to target $t$ at time step $k$ and $m^i_{k}$ is a measurement to a known landmark at time step $k$ for bias estimation. $x^t={[e^t,n^t]}^T$ is the east and north position of target $t\ =1,...,6$.
The tracking assignments for each agent, along with the measurements noise error covariances for the relative target ($R^{i,1}$) and landmark ($R^{i,2}$) measurements are given in Table \ref{tab:measurementError} and illustrated in Fig. \ref{fig:targetTrackingExample}.
The relative target measurement noise characteristics for different targets measured by the same agent are taken to be equal. 
For example, agent $1$ takes noisy measurements to targets $1$ and $2$ with $1\ m^2$ and $10\ m^2$ variances in the east and north directions, respectively, and $3\ m^2$ in both directions for the landmark. 
 
Following the definitions from Sec. \ref{sec:factorCF}, the full state vector includes 22 random states 
\begin{equation}
     \chi={[{(x^1)}^T,...,{(x^6)}^T,{(s^1)}^T,...,{(s^5)}^T]}^T,
     \label{eq:exampleFullState}
 \end{equation}
where for the HS-CF fusion, define the local random state vector at agent $i$
\begin{equation}
     \chi^i={[{(X^{\mathcal{T}_i})}^T,{(s^i)}^T]}^T.
     \label{eq:examplePartState}
\end{equation}
Here $\mathcal{T}_i$ is the set of targets observed by agent $i$ and  $X^{\mathcal{T}_i}$ includes all target random state vectors $x^t$, s.t $t\in \mathcal{T}_i$. 
In other words, the local random state vector at each agent includes only locally relevant targets and the local biases. 
In the HS-CF, where two agents $i$ and $j$ only share the marginal statistics regarding common states, messages should only consist data regarding targets $t\in \mathcal{T}_i \cap \mathcal{T}_j$. 
For example, according to Table \ref{tab:measurementError} and the network tree topology, for agents $1$ and $2$: $\mathcal{T}_1 \cap \mathcal{T}_2=T_2$, i.e. the common random state is $x^2$. Notice that the local subset of rvs is not limited to only one type of state. For example, agents $1$ and $5$ has target position state vectors $x^1$ and $x^6$, respectively, in their local subset, since they are the only agents tasked with the corresponding targets. 

The data communication requirements for this relatively small system were calculated: similar to the results from Sec. \ref{sec:NumExample}, the BDF-CF and the HS-CF require about 38\% and 2.6\% of the original CF communication data requirements, respectively.
\begin{comment}

\begin{table}[tb]
\renewcommand{\arraystretch}{1.2}
\caption{Local platform target assignments and sensor measurement error covariances. }
    \begin{center}
    \begin{tabular}{c|c|c|c}
        Agent    & Tracked Targets & $R^{i,1} [m^2]$ & $R^{i,2} [m^2]$  \\ \hline
        1 & $T_1,T_2$ & diag([1,10]) & diag([3,3]) \\ \hline
        2 & $T_2,T_3$ & diag([3,3]) & diag([3,3]) \\ \hline
        3 & $T_3,T_4,T_5$ & diag([4,4]) & diag([2,2]) \\ \hline
        4 & $T_4,T_5$ & diag([10,1]) & diag([4,4]) \\ \hline
        5 & $T_5,T_6$ & diag([2,2]) & diag([5,5]) \\ \hline
    \end{tabular}
    \end{center}
    \label{tab:measurementError}
    %\vspace{-0.22in}
\end{table} 
\end{comment}

\begin{table}[tb]
\renewcommand{\arraystretch}{1.2}
\caption{Local platform target assignments, common and local rv sets and sensor measurement error covariances. }
    \begin{center}
    \begin{tabular}{c|c|c|c|c|c}
        Agent    & Targets&$\chi_C^i$&$\chi_L^i$ & $R^{i,1} [m^2]$ & $R^{i,2} [m^2]$  \\ \hline
        1 & $T_1,T_2$& $x^2$ &$s^1, x^1$ &diag[1,10] & diag[3,3] \\ \hline
        2 & $T_2,T_3$& $x^2, x^3$& $s^2$&diag[3,3] & diag[3,3] \\ \hline
        3 & $T_3,T_4,T_5$ &$x^3, x^4, x^5$ & $s^3$ &diag[4,4] & diag[2,2] \\ \hline
        4 & $T_4,T_5$& $x^4, x^5$& $s^4$&diag[10,1] & diag[4,4] \\ \hline
        5 & $T_5,T_6$& $x^5$ & $s^5, x^6$&diag[2,2] & diag[5,5] \\ \hline
    \end{tabular}
    \end{center}
    \label{tab:measurementError}
    \vspace{-0.1in}
\end{table} 

The BDF-CF and the HS-CF performance was tested with 500 Monte Carlo simulations and compared to a centralized estimator. 
As mentioned before, in the BDF-CF each platform processes the full random state vector (\ref{eq:exampleFullState}), while in the HS-CF each platform processes only the locally relevant random states (\ref{eq:examplePartState}). 
In the simulations fusion occurs in every time step.

% Old figure
\begin{comment}
\begin{figure*}[bt!]
	\centering
    %\includesvg[width=0.7\textwidth]{Figures/static_50MC_a3_a4.svg}
     \includegraphics[width=0.7\textwidth]{Figures/static_50MC_a3_a4.pdf}
	\caption{Example 1 (static) - comparison between different fusion methods of agents 3 and 4 mean RMSE in full line. The dashed line shows the $2\sigma$ confidence bounds and the dotted lines present differences in $2\sigma$ between the CF derivatives and the centralized estimate. Results are shown in logarithm scale.  } 
	\label{fig:simResults4}
	%\vspace{-0.25in}
\end{figure*}
\end{comment}

\begin{figure}[bt!]
	\centering
    %\includesvg[width=0.7\textwidth]{Figures/static_50MC_a3_a4.svg}
     \includegraphics[width=0.48\textwidth]{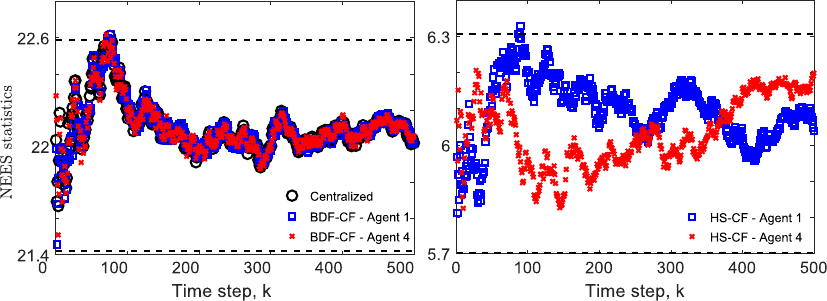}
	\caption{Example 1 (static) - NEES chi-square test based on 500 Monte Carlo simulations, where the dashed lines show bounds for 95$\%$ confidence level. Shown are test results of agents 1 and 4 using for different fusion methods, the results indicate all methods produce a consistent estimate.   }
	\label{fig:simResults4}
	\vspace{-0.2in}
\end{figure}

Fig. \ref{fig:simResults4} shows a NEES chi-square consistency test \cite{rong_practical_2001}, \cite{bar-shalom_linear_2001} results for agents 1 and 4 and a centralized estimator. 
Results are based on 500 Monte Carlo simulations with 95\% confidence level.
It can be seen that the MMSE estimates of both agents, with the BDF-CF and the HS-CF are consistent.
Note that since the HS-CF only estimates a subset of 6 states out of the full 22 state vector, the consistency bounds are different for these methods. The second test to determine the performance of a fusion method is whether it is conservative relative to a centralized estimator (see appendix \ref{sec:consist}). 
To validate, the local covariance matrix must be checked to see whether it is pessimistic relative to the centralized covariance. One simple test is by checking if the eigenvalues of $\bar{\Sigma}_{\chi^i}-\bar{\Sigma}_{\chi^i}^{cent}$ are equal or larger than zero, where $\bar{\Sigma}_{\chi^i}$ is the agent's covariance and $\bar{\Sigma}_{\chi^i}^{cent}$ is the centralized marginal covariance over $\chi^i$. 
% If the minimal eigenvalue is bigger or equal to zero, all eigenvalues are bigger or equal to zero and the MMSE estimate is conservative in the PSD sense. 
In the above simulations the minimal eigenvalues between all agents and all simulations were 0, for both the BDF-CF and the HS-CF, thus they are conservative in the PSD sense.

%\begin{figure*}[tb]
%	\centering
%	\includegraphics[width=0.8\textwidth]{Figures/A5_100MC_RMSE_wBDF_CF.eps}
%	\caption{RMSE - Agents 5}
%	\label{fig:simResults5}
%\end{figure*}

\subsection{Example 2 - Dynamic Case}
\label{sec:DynamicSim}
In dynamic systems, as discussed in Sec. \ref{sec:DynamicSystems}, there is a challenge in maintaining conditional independence. 
Two ways to solution are suggested, the first using the \emph{iAS}, thus keeping a distribution over the full time history over target random states, which is costly in both communication and computation requirements. 
The second, more efficient solution, is to perform conservative filtering by enforcing conditional independence in the marginalization step and deflating the information matrix (Algorithm \ref{algo:CF2}). 
Since this process loses information due to deflation, the BDF-CF becomes an approximate solution and is expected to be less accurate than the \emph{iAS} implementation. 
This is shown using a two agent, one (dynamic) target tracking simulation. 
Here the target follows a linear dynamics model with time-varying acceleration control,
\begin{equation}
    x_{k+1}=Fx_{k}+Gu_k+\omega_k, \ \ \omega_k \sim \mathcal{N}(0,0.08\cdot I_{n_x\times n_x}),
    \label{eq:dynamicEq}
\end{equation}
where
\begin{equation}
    \begin{split}
        F=\begin{bmatrix}1 & \Delta t &0 &0\\0 &1 &0 &0\\ 0 &0 &1 & \Delta t\\0& 0 &0 &1 \end{bmatrix}, \quad
        G=\begin{bmatrix}\frac{1}{2}\Delta t^2 &0\\\Delta t&0\\0 &\frac{1}{2}\Delta t^2\\0 &\Delta t \end{bmatrix}.
        %\quad u_k^j=\begin{bmatrix} u^{e,j} \\ u^{n,j} \end{bmatrix}_k
    \end{split}
    \label{eq:dynamicDef}
\end{equation}
The acceleration input in the east and north directions is given by $u_k=\begin{bmatrix} a_e\cdot \cos(d_e\cdot k \Delta t) \ \ a_n\cdot \sin(d_n\cdot k \Delta t) \end{bmatrix}^T$, where $a_e/a_n$ and $d_e/d_n$ define the east and north amplitude and frequency, respectively. The measurement model is as in the static example, given in (\ref{eq:measModel}) with noise parameters defined for agents 1 and 2 in Table \ref{tab:measurementError}.

\begin{figure}[tb]
	\centering
	\includegraphics[width=0.48\textwidth]{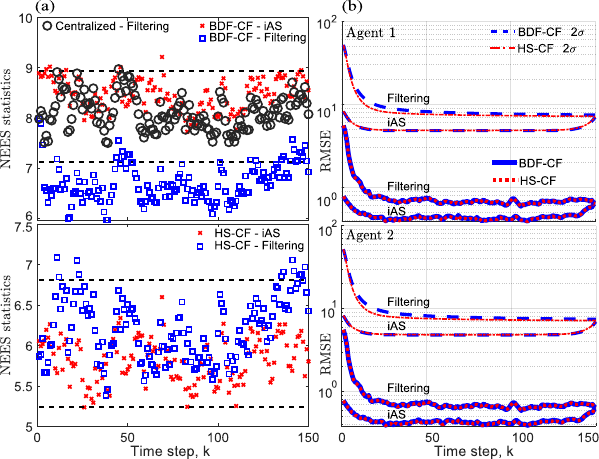}
	\caption{Results from a 75 Monte Carlo simulation of a 2 agent, 1 target dynamic target tracking scenario. Shown is a comparison between \emph{iAS} and filtering. (a) NEES chi-square test with the upper figure showing the BDF-CF compared to the centralized (with filtering) and the lower showing the HS-CF. Here the dashed black lines show bounds for 95$\%$ confidence level. (b) RMSE comparisons between the BDF-CF and the HS-CF for \emph{iAS} and filtering approaches for agent 1 (upper) and agent 2 (lower). }
	\label{fig:comp_BDF_iAS_marg}
	\vspace{-0.2in}
\end{figure}

Results in Fig. \ref{fig:comp_BDF_iAS_marg} show a comparison between the \emph{iAS} with the full time window ($k:1$) and filtering approaches (sliding window of size 1), using the BDF-CF and the HS-CF for fusion. 
The plots in figure (a) show the NEES chi-square consistency tests (75 simulations, 95\% confidence level) for agent 1, where the centralized (filtering results presented) and BDF-CF in the upper plot and the HS-CF with its different bounds in the lower.  %We can see that 
The results are consistent, with pessimistic behaviour of the BDF-CF due to the conservative filtering approach. 
(b) Shows the root mean squared error (RMSE) results of the same simulation, with agent 1 in the upper plot and agent 2 in the lower. %Here we can see that 
The \emph{iAS} has better performance, with smaller RMSE and $2\sigma$ bounds, which is to be expected due to its smoothing operation, but at the expense of much higher computation and communication load.    

The conservativeness of the fused estimate was checked again by computing the minimal eigenvalues across 75 simulations and the two agents.
For the \emph{iAS} approach, the BDF-CF and the HS-CF had small negative minimal eigenvalues of $-0.002$ and $-0.0015$ respectively, thus slightly overconfident relative to the centralized. 
For the conservative filtering approach, the BDF-CF was conservative with minimal eigenvalue of $0.0008$ and the HS-CF was overconfident with minimal eigenvalue of $-0.26$.

% old version - first TRO submission
\begin{comment}

\begin{figure}[tb]
	\centering
	\includegraphics[width=0.47\textwidth]{Figures/Dynamic_20MC_2A_1T_filtering_AS_BDF_comp_new.pdf}
	\caption{Results from a 20 Monte Carlo simulation of a dynamic target tracking scenario. Shown is a comparison between the iAS results with and without marginalization. The solid lines show the RMSE over target and agent ownship states relevant to that agent. The corresponding dashed lines show $2\sigma$ confidence bounds.} 
	\label{fig:comp_BDF_iAS_marg}
	\vspace{-0.2in}
\end{figure}
\end{comment}

Conservative filtering allows to test the algorithms on a more interesting dynamic simulation. 
As a test case, a simulation of a cooperative target tracking task with 4 agents and 5 dynamic targets (full random state vector of 28 states) was performed. 
The dynamic model details are the same as in the 2 agent, 1 target dynamic simulation above, with measurement parameters defined by the first 4 agents in Table \ref{tab:measurementError}. 
The advantages of the BDF-CF and HS-CF regarding communication and computation costs are highlighted again, as the BDF-CF saves 58\% in communication costs relative to the original CF, and the HS-CF saves 94.5\% in communication and 87.5\% in computation complexity.  

\begin{figure}[tb!]
	\centering
	\includegraphics[width=0.48\textwidth]{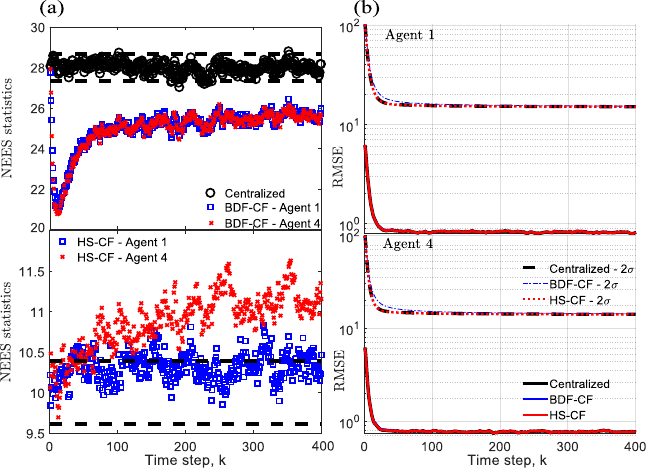}
	\caption{Results from a 500 Monte Carlo simulation of a cooperative target tracking task, with filtering, consisting of 4 tracking agents and 5 dynamic targets. Presented are results for agents 1 and 4. (a) NEES chi-square consistency test results for the centralized and BDF-CF estimates of 28 random states (upper) and the HS-CF estimates over 10 random states (lower).
	(b) Solid lines show the RMSE over target and agent local states relevant to that agent, dashed line shows the $2\sigma$ confidence bounds on a logarithmic scale, for agent 1 (upper) and 4 (lower).   } 
	\label{fig:dynanicSim}
	\vspace{-0.2in}
\end{figure}

Results of 500 Monte Carlo simulation with filtering for agents 1 and 4 are presented in Fig. \ref{fig:dynanicSim} (similar results for agents 2-3 are not presented to avoid cluttering the figure).  
The plots in (a) show the NEES chi-square consistency test with 95\% confidence bound marked with black dashed lines. The upper plot shows the centralized (black circles) and the BDF-CF for agents 1 (blue squares) and 4 (red x) NEES statistics for the 28-state vector. 
The lower plot shows the HS-CF results, which has a smaller 10-state random vector. 
(b) shows the corresponding RMSE for agent 1 (upper plot) and 4 (lower plot). 
Note that the RMSE results for the centralized estimate and BDF-CF, which hold distributions over the full 28-state vector, are marginalized and computed only over relevant local 10 agent random states for this comparison. 

It is seen from the NEES statistics plots that, as expected, the centralized estimator produces a consistent MMSE estimate, and the BDF-CF overestimates the uncertainty due to the information matrix deflation (covariance inflation) in the conservative filtering step. 
The BDF-CF also produces a conservative MMSE estimate relative to the centralized in the PSD sense for all agents, since the minimum eigenvalue between the agents is positive ($3e-04$). 
The HS-CF is slightly overconfident for both the consistency test and the PSD test, with negative minimal eigenvalue of $-0.26$. 
However, the degree of non-conservativeness in the HS-CF will in general be highly problem- and topology dependent.
Hence, the choice of whether to task agents with the full random state vector, with either homogeneous DDF methods (e.g., classical CF and conventional CI) or heterogeneous fusion with the BDF-CF, or to task them with only a subset of relevant random states using the HS-CF, will hinge on the desired trade-off in communication/computation complexity vs. resulting overconfidence in state MMSE estimates, provided that the HS-CF allows for stable convergence.        

The HS-CF overconfidence is attributed to inaccurate removal of implicit and hidden correlations due to marginalization in the filtering step (line 8 in Algorithm \ref{algo:CF2}). 
Correctly accounting for these dependencies is not in the scope of this paper, but is the focus of ongoing work \cite{dagan_conservative_2022}. 

% old figure - first version
\begin{comment}
\begin{figure*}[tb!]
	\centering
	\includegraphics[width=0.7\textwidth]{Figures/Dynamic_50MC_4A_5T_a3_a4.pdf}
	\caption{Results from a 50 Monte Carlo simulation of a cooperative target tracking task, with filtering, consisting 4 tracking agents and 5 dynamic targets. Presented are results for agents 3 and 4. Solid lines show the RMSE over target and agent ownship states relevant to that agent, dashed line shows the $2\sigma$ confidence bounds and the dotted lines present differences in $2\sigma$ between the BDF-CF and the centralized estimate. } 
	\label{fig:dynanicSim}
	\vspace{-0.2in}
\end{figure*}
\end{comment}

\section{Conclusions}
\label{sec:conclusions}
Heterogeneous fusion defines a key family of problems for Bayesian DDF, as it enables flexibility for large scale autonomous sensing networks. 
As shown in this work, separating the global joint distribution into smaller subsets of local distributions significantly reduces local agent computation and communication requirements. 
The heterogeneous fusion rules, analysis and derivations presented in this paper, assume acyclic networks for the purposes of exact fusion via the BDF- and HS-fusion algorithms, thus offer a basis for developing and analyzing similar algorithms. These assumptions can be gradually relaxed to solve more general heterogeneous scenarios involving exact or approximate fusion in more complex networked fusion settings.

Probabilistic graphical models (PGMs) were used here to develop Bayesian DDF algorithms. 
PGMs provided insight into the origin of the dependencies between random states not mutually tracked by two agents and enabled exploitation of the conditional independence structure embedded in these graphs. 
This led to two novel heterogeneous fusion rules for general probabilistic and Gaussian pdfs that were demonstrated on (but not limited to) static and dynamic target tracking problems. 
The latter motivated the development and use of the linear information augmented state (\emph{iAS}) smoother to regain conditional independence, on the expense of increasing computation and communication costs. 
To overcome this problem a conservative filtering approach was demonstrated to maintain conditional independence over a small time window, without the need of the full time history. 

The DDF framework naturally enables sparse distributed estimation for high-dimensional state estimation and the heterogeneous fusion rules represent a practical and theoretical shift in the state of the art, subject to usual provisos and limitations of DDF.
From a practical standpoint, the algorithms developed in this paper can already be used to improve scalability in a variety of decentralized Bayesian estimation problems, such as cooperative localization and navigation \cite{dourmashkin_gps-limited_2018,loefgren_scalable_2019}, multi-agent SLAM \cite{cunningham_ddf-sam_2013} and terrain height mapping \cite{schoenberg_distributed_2009}, where a height distribution is estimated on a grid map. In this case, for example, the BDF-CF can be used to reduce communication in the network by dividing the the map into several overlapping regions of interest, allowing agents to communicate only regarding those cells in which they have new data to contribute. This scales the communication with the number of locally observed grid cells instead of the entire map.

Indeed, some works are already leveraging heterogeneous DDF ideas for robotics, \cite{loefgren_scalable_2019, whitacre_decentralized_2011}, despite the gap in theoretical guarantees and understanding on the full nature of the problem and its limitations. 
This paper makes progress by building theoretical foundations for future research and surfacing a discussion on the assumptions and definitions of homogeneous DDF, as they appear to be inadequate for real world robotics problems of heterogeneous fusion. %We believe that 
Heterogeneous fusion, as defined in this paper, requires a careful revisit of the idea of `ideal' centralized/decentralized Bayesian estimation as well as the definitions of consistency and conservativeness for general (non-Gaussian) pdfs and more specifically in the case of heterogeneous pdfs in dynamic systems. 

%Other work will focus on two main avenues: i) theoretical consistency and conservativeness analysis of the proposed CF$^2$ methods; ii) development of heterogeneous fusion method for ad hoc networks with unknown common information distributions between fusing agents. 

\appendices
\section{What is A Good Fusion Rule?}
\label{sec:consist}
In the problem formulation (Sec. \ref{sec:ProbStatement}) it is left to define a good fusion rule $\mathbb{F}$  (\ref{eq:probStatement}) and how to evaluate it.
In a recent paper, Lubold and Taylor \cite{lubold_formal_2022} claim that a fusion rule should provide a posterior pdf which is conservative, i.e., ``overestimates the uncertainty of a system".
They suggest new definitions for conservativeness, but to the best of our knowledge it is not widely used. 
In the context of homogeneous fusion, common definitions use the terms consistent\footnote{Here the term consistent means `filter consistent' (see Bar-Shalom \cite{bar-shalom_linear_2001}).} and conservative interchangeably \cite{julier_non-divergent_1997}, \cite{uhlmann_covariance_2003} and assume that the uncertainty of a point estimate can be described by its mean and covariance. 
In the following, the intuition regarding conservativeness from \cite{lubold_formal_2021} is combined with the common definitions of consistency from homogeneous fusion to define a `good' heterogeneous fusion rule $\mathbb{F}$, firstly in terms of pdfs and then in the case of Bayesian point estimation.

From the standpoint of pdf fusion, a good heterogeneous fusion rule $\mathbb{F}$ results in an updated local posterior pdf that: (i) does not underestimate the uncertainty relative to the true pdf, where Bar-Shalom \emph{et al.} dub this `dynamic (filter) consistency' \cite{bar-shalom_linear_2001} and (ii) is conservative over the agent's random states of interest $\chi^i_{k}\subseteq \chi_k$ relative to a consistent centralized estimators' marginal pdf over $\chi^i_{k}$, 
\begin{equation}
     p^i(\chi^i_k)\succeq \int p^{cent}(\chi_k)d\chi^{\neg i}_k,
     \label{eq:marginalConservative}
\end{equation} 
where `$\succeq$' denotes conservative and $\chi^{\neg i}_k=\chi_k\setminus \chi^i_k$ is the set of variables not included in agent $i$'s random states of interest.

The centralized pdf refers to the posterior pdf over the full random state vector $\chi_k$ conditioned on all the available data from all the agents up and including time step $k$,  $p^{cent}(\chi_k|\bigcup_{i\in N_a} Z^{i,-}_{k})$.

Since consistency and conservativeness are often defined by the first two moments of the pdf, i.e., the mean and covariance, the above definition can be further narrowed in the context of Bayesian point estimation. 
A good heterogeneous fusion rule $\mathbb{F}$ in this case is then one that when forming a point estimate from its resulting local posterior $p^i(\chi^i_{k}|Z_{k}^{i,+})$, for
example by finding the minimum mean squared error (MMSE) estimate, the estimate: (i) does not underestimate the uncertainty relative to the true state error statistics, and (ii) is conservative relative to the marginal error estimate of a consistent centralized point estimator.

For example, assume the means of the Gaussian random state vectors $\chi^i$ and $\chi$ are $\mu_{\chi^i}$ and $\mu_{\chi}$, and the covariances, describing the mean squared error, are $\Sigma_{\chi^i}=E[(\chi^i-\mu_{\chi^i})(\chi^i-\mu_{\chi^i})^T]$ and $\Sigma_{\chi}=E[(\chi-\mu_{\chi})(\chi-\mu_{\chi})^T]$, respectively, where $E[\cdot]$ is the expectation operator.
The actual values are unknown, and the approximate estimate of them is given by $\bar{\mu}_{\chi^i}$,  $\bar{\mu}_{\chi}$, and $\bar{\Sigma}_{\chi^i}$, $\bar{\Sigma}_{\chi}$.\\
The definitions above then translate to the following:
\begin{enumerate}
    \item Not underestimating of the uncertainty relative to the true error statistics implies that $\bar{\Sigma}_{\chi^i}-\Sigma_{\chi^i} \succeq 0$, i.e., the resulting matrix difference is PSD.
    \item Conservativeness relative to the marginal estimate of the centralized estimator implies that $\bar{\Sigma}_{\chi^i}-\bar{\Sigma}_{\chi^i}^{cent} \succeq 0$, where $\bar{\Sigma}_{\chi^i}^{cent}$ is the marginal covariance over $\chi^i$, taken from the joint centralized covariance over $\chi$ (\ref{eq:marginalConservative}).
\end{enumerate}
The centralized estimate in this case can be considered consistent if, for example, it passes the NEES chi-square test \cite{rong_practical_2001},  \cite{bar-shalom_linear_2001}. Note that since a consistent centralized estimate neither overestimates nor underestimates the uncertainty, a conservative (higher uncertainty) local estimate is expected to not underestimate the uncertainty. Thus, requiring the local estimate to be conservative relative to a consistent centralized estimate implicitly requires it to not underestimate the uncertainty of the true error statistics.

\ifCLASSOPTIONcaptionsoff
  \newpage
\fi



%\bibliographystyle{IEEEtran}
%\begin{thebibliography}{1}

\bibliographystyle{IEEEtran}
\bibliography{references.bib}
%\end{thebibliography}


\enlargethispage{-2.8in}
\begin{IEEEbiography}[{\includegraphics[width=1.1in,height=1.25in,clip,keepaspectratio]{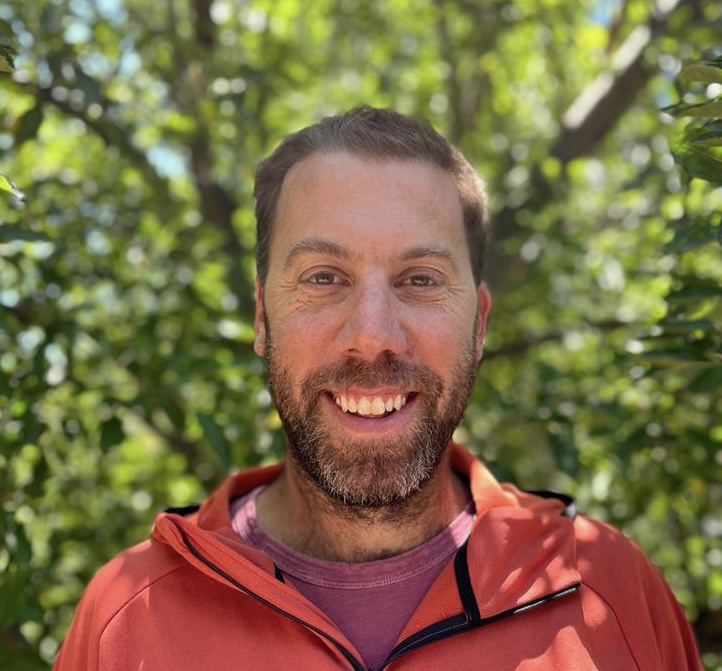}}]{Ofer Dagan}
received the B.S. degree in aerospace engineering, in 2010, and the M.S. degree in mechanical engineering, in 2015, from the Technion - Israel Institute of Technology, Haifa, Israel. 
He is currently working toward the Ph.D.
degree in aerospace engineering with the Ann and H.J.
Smead Aerospace Engineering Sciences Department,
University of Colorado Boulder, Boulder, CO, USA. 
From 2010 to 2018 he was a research engineer in the aerospace industry. 
His research interests include theory and algorithms for decentralized Bayesian reasoning in heterogeneous autonomous systems.
\end{IEEEbiography}

\begin{IEEEbiography}[{\includegraphics[width=1.1in,height=1.25in,clip,keepaspectratio]{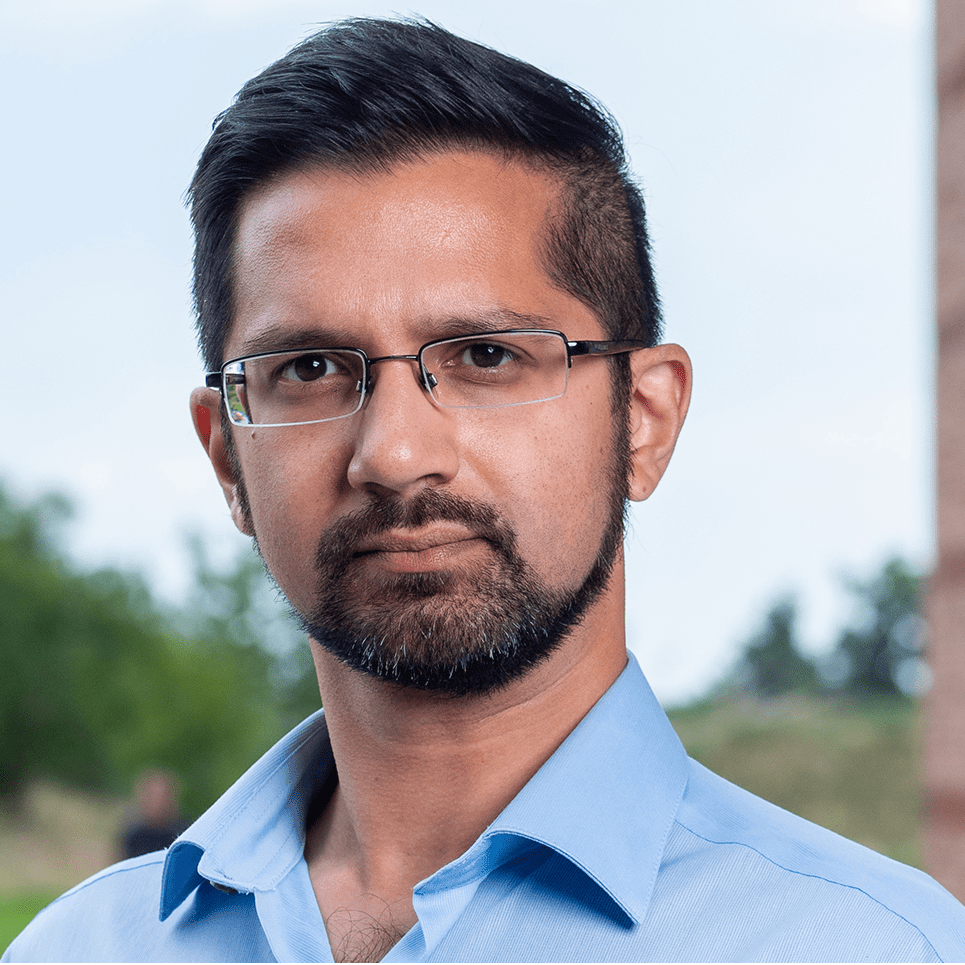}}]{Nisar R. Ahmed}
received the B.S. degree in engineering from Cooper Union, New York City, NY,USA, in 2006 and the Ph.D. degree in mechanical engineering from Cornell University, Ithaca, NY, USA,
in 2012.
He is an Associate Professor of Autonomous Systems and H. Joseph Smead Faculty Fellow with Ann and H.J. Smead Aerospace Engineering Sciences Department, University of Colorado Boulder, Boulder, CO, USA. He was also a Postdoctoral Research Associate with Cornell University
until 2014. His research interests include the development of probabilistic models and algorithms for cooperative intelligence in mixed human–machine teams.
\end{IEEEbiography}

%\newpage



%\vfill


\end{document}